\documentclass[
reprint,
superscriptaddress,
 amsmath,
 amssymb,
 aps,
 prc,
 floatfix,
]{revtex4-2}
\usepackage[utf8]{inputenc}
\usepackage[T1]{fontenc}
\usepackage{braket}
\usepackage{graphicx}
\usepackage{dcolumn}
\usepackage{bm}
\usepackage{xcolor}
\usepackage{mathtools}
\usepackage{soul}
\setstcolor{red}
\usepackage[caption=false]{subfig}
\usepackage{tikz}
\usepackage{graphicx}
\usepackage{booktabs}
\usepackage{multirow}
\usepackage{adjustbox}
\usepackage{comment}

\definecolor{JeffersonBlue}{RGB}{35, 45, 75} 
\definecolor{OxfordBlue}{rgb}{0,0.106,0.329}
\definecolor{GoldDecoration}{RGB}{170, 120, 70}

\usepackage[colorlinks=true,linkcolor=OxfordBlue!90,citecolor=OxfordBlue!90,urlcolor=OxfordBlue!90]{hyperref}
\newcolumntype{M}[1]{>{\centering\arraybackslash}m{#1}}

\begin{document}

\title{Compton Form Factor Extraction using Quantum Deep Neural Networks}

\author{Brandon B. Le}
 \email{sxh3qf@virginia.edu}
 \author{D. Keller}
 \email{dustin@virginia.edu}
\affiliation{%
 Department of Physics, University of Virginia, Charlottesville, VA 22904, USA.
}%

\date{\today}

\begin{abstract}
We extract Compton form factors (CFFs) from deeply virtual Compton scattering measurements at the Thomas Jefferson National Accelerator Facility (JLab) using quantum‑inspired deep neural networks (QDNNs). The analysis implements the twist‑2 Belitsky–Kirchner–Müller formalism and employs a fitting strategy that emulates standard local fits. Using pseudodata, we benchmark QDNNs against classical deep neural networks (CDNNs) and find that QDNNs often deliver higher predictive accuracy and tighter uncertainties at comparable model complexity. Guided by these results, we introduce a quantitative selection metric that indicates when QDNNs or CDNNs are optimal for a given experimental fit. After obtaining local extractions from the JLab data, we perform a standard neural‑network global CFF fit and compare with previous global analyses. The results support QDNNs as an efficient and complementary tool to CDNNs for CFF determination and for future multidimensional studies of parton distributions and hadronic structure.
\end{abstract}

\maketitle


\section{\label{sec:introduction}Introduction}
Mapping the quark and gluon structure of hadrons remains a central objective in QCD phenomenology. While traditional determinations of parton distribution functions (PDFs) have established the longitudinal momentum structure of the nucleon, a multidimensional description correlating longitudinal momentum with transverse spatial degrees of freedom is a major focus of current global facilities and the forthcoming Electron–Ion Collider (EIC)~\cite{Mueller1994,Ji1_97,Ji2_97,Rad_96,Rad_97}. Generalized Parton Distributions (GPDs) provide such a unified framework by interpolating between PDFs and elastic form factors, thereby encoding correlations between intrinsic longitudinal motion and transverse spatial structure. Through Ji’s sum rule, GPDs are directly connected to the total angular momentum carried by quarks and gluons, and through their relation to matrix elements of the QCD energy–momentum tensor, they provide access to gravitational form factors and the internal mechanical properties of hadrons, including pressure and shear distributions~\cite{spinpuzzle,Burkardt,Diehl,Dupre,POLYAKOV1,POLYAKOV2}. Comprehensive discussions of these mechanical interpretations and gravitational form factors can be found in Refs.~\cite{Burkert:2023RMPGFF,Polyakov:2018zvc,JiYang:2025PressureForces}.

Experimentally, GPDs are accessed in hard exclusive reactions, notably deeply virtual Compton scattering (DVCS), deeply virtual meson production (DVMP), time‑like Compton scattering (TCS), and double deeply virtual Compton scattering (DDVCS)~\cite{Ji2_97,Rad_96,BergerTCS,Guidal_DDVCS}. At leading twist the nucleon structure is encoded in four chiral‑even and four chiral‑odd quark GPDs (with analogous gluon GPDs) as functions of $(x,\xi,t)$, where $x\pm\xi$ denote the struck‑parton momentum fractions and $t$ is the momentum transfer; the physical interpretation separates into DGLAP and ERBL regions~\cite{Diehl_GPDs,Hoodbhoy_GPDs,Dokshitzer:1977sg,Gribov:1972ri,Lipatov:1974qm,Altarelli:1977zs,Efremov:1979qk,Lepage:1980fj}. QCD factorization theorems establish that the DVCS amplitude factorizes into a hard perturbative kernel and nonperturbative GPDs in the Bjorken limit~\cite{JCCollins_Fact,JiOs1997}. Because $x$ is not directly observable, phenomenology proceeds through Compton Form Factors (CFFs), which are convolutions of GPDs with calculable coefficient functions at leading order (LO) and beyond~\cite{Belitsky2002}. The real and imaginary parts of the CFFs probe principal‑value integrals and the lines $x=\pm\xi$, respectively, and their recovery from data constitutes an ill‑posed deconvolution problem~\cite{Deconv1,Deconv2} with additional $Q^2$ dependence from QCD evolution~\cite{Mueller1994}.

Over the past two decades, a broad experimental program has mapped DVCS over complementary kinematics. H1 and ZEUS at HERA measured unpolarized cross sections and beam‑charge asymmetries at small $x_B$~\cite{Aktas2005,Aaron2009,Chekanov2009}; HERMES provided the most complete fixed‑target set in the intermediate region~\cite{Airapetian2001,Airapetian2007,Airapetian2008,Airapetian2009,Airapetian2010,Airapetian2011,Airapetian2012}; at Jefferson Lab, CLAS and Hall~A produced finely binned cross sections and polarization observables at large $x_B$ in the valence region, including neutron DVCS for flavor separation~\cite{Stepanyan2001,Camacho2006,JoCLAS:2015,Defurne2015,Defurne2017,Georges2022,HallAn}; and COMPASS has reported cross sections with fixed targets at lower $x_B$~\cite{Akhunzyanov2019} and more recently \cite{compass2025measurement} for measurement of the hard exclusive $\pi^0$ production cross section. New data at JLab12 (CLAS12) and future measurements at AMBER, EIC, EicC, and LHeC will greatly extend coverage and precision~\cite{Gautheron2010,Biselli2006,clas12,EIC1,EIC2,EicC,LHeC}.

CFF extractions have followed two complementary strategies. \emph{Global} fits model the $x$‑dependence via GPD ans\"atze (e.g., VGG, GK, KM) or parameterize CFFs over all kinematics, enabling interpolation/extrapolation but inheriting model bias~\cite{KM09,Kumericki2015,kumericki:2020,GK,VGG,Kumericki:DR3}. \emph{Local} fits determine CFFs pointwise in $(x_B,t,Q^2)$ with minimal functional assumptions, at the cost of limited predictive power~\cite{Guidal:fit08,Guidal-Mou:fitHermes09,Guidal:fitCLAS10,Guidal:Htilde10,BoerGuidal:fit14,Kumericki:fitHermes14,Moutarde:fit09,Dupre,Georges2022}. To reduce modeling bias while retaining flexibility, neural‑network approaches have been developed for GPDs/CFFs that respect theory constraints and propagate experimental uncertainties through replica methods~\cite{KMS11,Moutarde:ANN2019,kumericki:2020,GrigsbyKriesten:DNN,Ripley87}; see Ref.~\cite{Kumericki:review2016} for a review. First‑principles information from lattice QCD is beginning to provide additional constraints~\cite{LaMET1,LaMET2,LaMET3,Guo2025PRLGUMP}.

In this work we explore whether quantum-inspired deep neural networks (QDNNs)—simulated hybrid quantum-classical architectures—can improve CFF extractions under realistic conditions. QDNNs employ parameterized quantum circuits as trainable layers in hybrid quantum–classical models, offering expressive Hilbert‑space feature maps and entanglement that may capture interference patterns and nontrivial correlations less efficiently represented by classical networks~\cite{Beer2020}. We implement QDNNs using PennyLane’s differentiable quantum framework, enabling end‑to‑end training with classical optimizers~\cite{bergholm2018pennylane}. To enable a like‑for‑like comparison, both the classical DNN (CDNN) and QDNN are trained within the same helicity‑amplitude formalism—BKM10 at twist‑2—and use identical physics‑informed losses built from the DVCS/BH decomposition~\cite{Belitsky2010}.

We adopt the unpolarized twist-2 framework used previously Ref.~\cite{Diaz:2025}, in which the
$\phi$-dependence is driven by the BH--DVCS interference through
$\Re e\,\mathcal H$, $\Re e\,\mathcal E$, and $\Re e\,\widetilde{\mathcal H}$,
together with a $\phi$-independent DVCS contribution. We benchmark a CDNN and a QDNN on
Monte Carlo pseudodata replicas, introduce a data-driven \emph{DVCS quantum
qualifier} to select \emph{a priori} the better model class in each kinematic bin based
on experimental uncertainty and the nonlinearity of the $\phi$-dependent signal,
apply the qualifier-guided QDNN extraction to the JLab cross section data and
compare with the KM15 global analysis~\cite{Kumericki2015}, and finally construct a global
parameterization by training a separate DNN on the combined set of best-bin
CDNN/QDNN local extractions.

Overall, this study benchmarks a quantum-circuit neural-network approach for CFF
extraction in the sparse, noisy regime typical of DVCS measurements. The QDNN is
implemented on classical hardware via a \emph{quantum simulator}, using a
parameterized quantum circuit as an alternative model class—and thus an
alternative algorithmic route to inference—relative to a standard CDNN. We do
not require a physical quantum device; using one would introduce additional
device-induced uncertainties from decoherence and hardware noise. Within a common
BKM10-based forward model and training procedure, we present both separate and
combined CDNN/QDNN extractions from experimental cross sections, obtaining
substantially reduced uncertainties on $\Re e\,\mathcal H$, $\Re e\,\mathcal E$,
and $\Re e\,\widetilde{\mathcal H}$ relative to representative previous
extractions in the literature.

The remainder of this paper is organized as follows. In Sec.~\ref{sec:TheoryFramework} we summarize the unpolarized BKM10 twist-2 formalism and define the forward model for the DVCS+BH cross section that enters the physics-informed loss. Section~\ref{sec:QuantumDNNs} introduces the CDNN and QDNN model classes, our quantum-circuit design choices, and the common training procedure implemented on a quantum simulator. In Sec.~\ref{sec:ExperimentalData} we describe the JLab datasets, kinematic selections, and the pseudodata/replica construction used for closure tests. Section~\ref{sec:CFFExtraction} presents like-for-like benchmarks of CDNN and QDNN extractions on pseudodata, including comparisons between the baseline and fully optimized quantum architectures and a breakdown of precision, accuracy, and algorithmic contributions. In Sec.~\ref{sec:dvcs-qq} we construct and validate the DVCS quantum qualifier used to predict, \emph{a priori}, which model class is expected to perform best in a given kinematic bin. Section~\ref{sec:cff-extract-exp-data} applies the qualifier-guided strategy to experimental cross sections, produces local CFF extractions, and uses the combined best-bin results to build a global DNN parameterization for comparison with established analyses. We summarize our main findings and outline extensions to broader observable sets and increased theory fidelity in Sec.~\ref{sec:con}.

\section{\label{sec:TheoryFramework}Theory Framework}

This section defines the photon electroproduction cross section and specifies the
twist-2 Belitsky--Kirchner--M\"uller (BKM10) helicity-amplitude formalism used
throughout the paper to compute the model prediction $F_{\rm pred}(\phi)$ that
enters the physics-informed loss (Sec.~III and Eq.~(\ref{eq:loss})).  In the unpolarized
setting considered here, we treat the Bethe--Heitler (BH) contribution as known,
and we constrain the $\phi$-dependence of the cross section through the
BH--DVCS interference term at twist-2.  Consistent with the simplified BKM10
setup used in Ref.~\cite{Diaz:2025}, the only free quantities entering our
forward model are
$\Re e\,\mathcal{H}(x_B,Q^2,t)$,
$\Re e\,\mathcal{E}(x_B,Q^2,t)$,
$\Re e\,\widetilde{\mathcal{H}}(x_B,Q^2,t)$,
and $\mathrm{DVCS}(x_B,Q^2,t)$,
where $\mathrm{DVCS}$ parametrizes the (twist-2) $\phi$-independent DVCS-squared contribution (details below).

In the helicity-independent (unpolarized beam, unpolarized target) case, the
4-fold photon electroproduction differential cross section can be written as
\begin{eqnarray}
\frac{d^4\sigma}{d x_{B} d Q^2 d|t| d\phi}
= \frac{\alpha^3 x_B y^2}{8 \pi Q^4 \sqrt{1 + \epsilon^2}} \frac{1}{e^6}
\big|\mathcal{T}\big|^2 \;,
\end{eqnarray}
with the standard DIS/DVCS variables
$x_B = Q^2/(2\,p\!\cdot\! q)$, $q=k-k'$ and $Q^2 \equiv -q^2$,
$t=\Delta^2$ with $\Delta=p'-p$, and $y=(p \cdot q)/(p \cdot k)$.
The azimuthal angle $\phi$ between the leptonic and hadronic planes is defined
in the Trento convention~\cite{Bacchetta2004}.  We use
$\alpha=e^2/(4\pi)$ and $\epsilon^2 \equiv 4 x_B^2 M^2/Q^2$.

The measured final state is shared by deeply virtual Compton scattering (DVCS)
and the Bethe--Heitler (BH) process.  Experimentally the channels are
indistinguishable, so the cross section is governed by the coherent sum of
amplitudes,
\begin{equation}\label{eq:total_amp}
|\mathcal{T}|^2
= |\mathcal{T}_{DVCS}|^2 + |\mathcal{T}_{BH}|^2 + \mathcal{I},
\end{equation}
with interference term
\begin{equation}
\mathcal{I}
= \mathcal{T}_{DVCS}\,\mathcal{T}_{BH}^*
+ \mathcal{T}_{DVCS}^*\,\mathcal{T}_{BH}.
\end{equation}
In our extraction, Eq.~\eqref{eq:total_amp} is the starting point for the
\emph{forward model} used to compute $F_{\rm pred}$ in the loss.

The BH contribution is computed exactly in Ref.~\cite{Belitsky2002} and depends
only on QED kinematics and the elastic proton form factors.  In particular,
for the unpolarized case one may write
\begin{equation}
\resizebox{0.9\hsize}{!}{
$\big|\mathcal{T}_{BH}\big|^2
= \displaystyle\frac{e^6}{x_B^2 y^2(1 + \epsilon^2)^2\; t\;\mathcal{P}_1(\phi)\mathcal{P}_2(\phi)}
\sum\limits_{n=0}^{2} c^{BH}_n \cos(n\phi)$,}
\end{equation}
where $c^{BH}_n$ are functions of $F_1(t)$ and $F_2(t)$ (taken from Kelly's
parametrization~\cite{KellyParams}) and $\mathcal{P}_{1,2}(\phi)$ are the lepton
propagators.  \emph{Practically,} $|\mathcal{T}_{BH}|^2$ is treated as a known
piece of the forward model and introduces no fit parameters.

At leading twist, QCD factorization theorems~\cite{JiOs1997,white2001} imply that
DVCS factorizes into a perturbatively calculable hard coefficient function and
nonperturbative GPDs.  Because the struck-parton
momentum fraction $x$ is not directly observable, DVCS observables are expressed
in terms of CFFs, i.e., convolutions of GPDs with the hard
kernel.

For completeness and to fix conventions, at leading order the twist-2 quark CFFs
$\mathcal{F}=\{ \mathcal{H}, \mathcal{E}, \widetilde{\mathcal{H}}, \widetilde{\mathcal{E}}\}$
are related to the corresponding quark GPDs
$F = \{H,E,\widetilde{H}, \widetilde{E}\}$ through
\begin{equation} \label{eq:CFFs_conv}
\mathcal{F}(\xi,t)
\equiv \sum\limits_{q} \int_{-1}^{1} dx\,
C_q^{[\mp]}(x,\xi)\,F_q(x,\xi,t),
\end{equation}
with
\begin{equation}
C_q^{[\mp]}
= e_q^2\left(\frac{1}{\xi-x-i0}\mp\frac{1}{\xi+x-i0}\right),
\end{equation}
and $\xi=x_B/(2-x_B)$ at leading twist.
Each CFF is complex,
\begin{equation}
\mathcal{F}(\xi,t) = \Re e\,\mathcal{F}(\xi,t) + i\,\Im m\,\mathcal{F}(\xi,t).
\end{equation}
\emph{Importantly for this work:} we do \emph{not} attempt a GPD deconvolution or
evaluate Eq.~\eqref{eq:CFFs_conv} in the fit.  Instead, the relevant CFF
components are treated as effective, kinematic-dependent quantities to be
regressed from data via the physics-informed forward model defined below.

We work in the BKM10 formalism~\cite{Belitsky2010} and restrict to a leading-twist
regime by imposing $Q^{2} > 1.5~\mathrm{GeV}^{2}$ and $|t|/Q^{2} \le 0.25$
(consistent with a twist-2 treatment~\cite{kumericki:2020}).
Within this setup:

\begin{itemize}
\item The DVCS-squared contribution is $\phi$-independent at twist-2 for
unpolarized kinematics.  While the full expression can be written as a bilinear
combination of CFFs, in this paper we absorb the entire DVCS-squared term
into a single effective constant $\mathrm{DVCS}(x_B,Q^2,t)$ to be fitted.
\item The interference term provides the dominant $\phi$-dependence and,
under the helicity-conserving/twist-2 restriction adopted here, depends only on
$\Re e\,\mathcal{H}$, $\Re e\,\mathcal{E}$, and $\Re e\,\widetilde{\mathcal{H}}$.
\end{itemize}

Concretely, the unpolarized interference term has the harmonic structure
\begin{eqnarray}
\label{equ:Interference1}
\mathcal{I}
= \frac{e^6}{x_By^3 t\,\mathcal{P}_1(\phi)\mathcal{P}_2(\phi)}
\sum\limits_{n=0}^{3} c^{\mathcal{I}}_n \cos(n\phi),
\end{eqnarray}
with Fourier coefficients
\begin{align}
\label{equ:Interference2}
c^{\mathcal{I}}_n
&= C^n_{++}\,\Re e\, C^{\mathcal{I},n}_{++}(\mathcal{F})
\nonumber\\
&\quad + C^n_{0+}\,\Re e\, C^{\mathcal{I},n}_{0+}(\mathcal{F}_{\mathrm{eff}})
\nonumber\\
&\quad + C^n_{-+}\,\Re e\, C^{\mathcal{I},n}_{-+}(\mathcal{F}_T)\, .
\end{align}
We neglect the double helicity-flip gluonic amplitudes
$C^{\mathcal{I},n}_{-+}(\mathcal{F}_T)$ (formally suppressed by $\alpha_s$) and
we drop helicity-changing/twist-3 effects contained in the effective CFFs
$\mathcal{F}_{eff}$.  With this choice, only the helicity-conserving amplitudes
remain,
\begin{align}
\label{equ:Interference3}
\Re e\, C^{\mathcal{I},n}_{++}(\mathcal{F})
&= \Re e\, C^{\mathcal{I}}(\mathcal{F})
\nonumber\\
&\quad + \frac{C^{V,n}_{++}}{C^{n}_{++}}\,
\Re e\, C^{\mathcal{I},V}(\mathcal{F})
\nonumber\\
&\quad + \frac{C^{A,n}_{++}}{C^{n}_{++}}\,
\Re e\, C^{\mathcal{I},A}(\mathcal{F}) \, .
\end{align}
where the kinematic coefficients are given in Ref.~\cite{Belitsky2010}, and the
linear CFF combinations entering the interference are
\begin{eqnarray}
\label{equ:Interference4}
C^{\mathcal{I}}(\mathcal{F})
= F_1\,\mathcal{H}
+\xi(F_1+F_2)\,\widetilde{\mathcal{H}}
-\frac{t}{4M^2}\,F_2\,\mathcal{E},
\end{eqnarray}
\begin{eqnarray}
\label{equ:Interference5}
C^{\mathcal{I},V}(\mathcal{F})
= \frac{x_B}{2-x_B+x_B\frac{t}{Q^2}}(F_1+F_2)\,(\mathcal{H}+\mathcal{E}),
\end{eqnarray}
\begin{eqnarray}
\label{equ:Interference6}
C^{\mathcal{I},A}(\mathcal{F})
= \frac{x_B}{2-x_B+x_B\frac{t}{Q^2}}(F_1+F_2)\,\widetilde{\mathcal{H}}.
\end{eqnarray}

With the restrictions above, the $\phi$-dependence of the unpolarized cross
section is governed by the interference term through the three real CFFs
$\Re e\,\mathcal{H}$, $\Re e\,\mathcal{E}$, and $\Re e\,\widetilde{\mathcal{H}}$,
while the DVCS-squared contribution is represented by a single $\phi$-independent
parameter $\mathrm{DVCS}$.

To make the connection to the remainder of the paper explicit, the model
prediction entering the loss (Eq.~(\ref{eq:loss})) can be viewed schematically as
\begin{align}
\label{eq:forward_model}
F_{\rm pred}(x_B,Q^2,t,\phi)
&= F_{\rm BH}(x_B,Q^2,t,\phi)
\nonumber\\
&\quad + F_{\rm INT}\!\left(
x_B,Q^2,t,\phi;
\Re e\,\mathcal{H},
\Re e\,\mathcal{E},
\Re e\,\widetilde{\mathcal{H}}
\right)
\nonumber\\
&\quad + \mathrm{DVCS}(x_B,Q^2,t)\, .
\end{align}
where $F_{\rm BH}$ is computed from the known BH expression and form factors,
$F_{\rm INT}$ is computed from Eqs.~\eqref{equ:Interference1}--\eqref{equ:Interference6},
and $\mathrm{DVCS}$ is fitted.  This is the exact simplified BKM10 formalism
used in Ref.~\cite{Diaz:2025} and adopted here to enable a direct comparison of
CDNN and QDNN extractions within the same physics-informed forward model.

The inference strategy adopted in this work is forward-model agnostic: the physical content enters entirely through the DVCS+BH cross-section model $F_{\mathrm{pred}}$, while the CDNN or QDNN acts solely as a regressor and optimizer for the effective CFF parameters. At strict leading twist (twist-2) and leading power in $1/Q$, different formulations of the DVCS amplitude are equivalent descriptions of the same underlying physics once conventions (e.g., azimuthal angle definitions and CFF normalizations) are consistently aligned. In this limit, consistent implementations of different twist-2 formalisms are expected to yield compatible extracted CFF values up to trivial convention conversions.

Differences between published cross-section formulas typically arise from kinematical power corrections or convention-dependent treatments of subleading terms that effectively enter at twist-3 and beyond \cite{Guo:2021KinematicalDVCS}. These effects are discussed further in Sec.~\ref{sec:cff-extract-exp-data}.

\section{\label{sec:QuantumDNNs}Quantum Deep Neural Networks}

In this work, QDNNs are used as compact, trainable feature maps for the local regression of the three real CFFs and the $\phi$‑independent DVCS term at twist‑2,
\[
f_{\Theta}(x_B,Q^2,t)=\big(\Re e \mathcal{H},\ \Re e \mathcal{E},\ \Re e \mathcal{\widetilde{H}},\ \mathrm{DVCS}\big),
\]
which then enter the BKM10 helicity‑amplitude formalism of Sec.~\ref{sec:TheoryFramework} to predict the unpolarized cross section via Eq.~\ref{eq:total_amp}. The QDNNs are implemented with \textsc{PennyLane}~\cite{bergholm2018pennylane} and trained end‑to‑end on a state‑vector simulator, allowing a like‑for‑like benchmark against a CDNN under the \emph{same} physics‑informed loss (Sec.~\ref{sec:CFFExtraction}). Unless stated otherwise we use $n=6$ qubits (a $2^6$‑dimensional embedding; cf.\ Fig.~2).

Let $x\equiv(x_B,Q^2,t)$. Classical features are encoded by a fixed \emph{angle‑embedding} block $E(x)$ acting on $\ket{0}^{\otimes n}$, followed by $L$ trainable unitary blocks and a projective readout:
\begin{align}\nonumber
\ket{\psi_0(x)} &= E(x)\,\ket{0}^{\otimes n},\\\nonumber
\ket{\psi_L(\boldsymbol\theta,x)} &= U_L(\boldsymbol\theta_L)\cdots U_1(\boldsymbol\theta_1)\,\ket{\psi_0(x)},\\\nonumber
z_j(\boldsymbol\theta,x) &= \bra{\psi_L(\boldsymbol\theta,x)}\,Z_j\,\ket{\psi_L(\boldsymbol\theta,x)}\qquad (j=1,\dots,n).\label{eq:zreadout}
\end{align}
Each $U_\ell$ is a \emph{StronglyEntanglingLayer} (SEL) built from single‑qubit rotations and fixed two‑qubit entanglers with a tunable range; this balances expressivity and stability while keeping the number of trainable parameters modest. The measured expectations $z\in\mathbb{R}^n$ feed a small classical head $g_{\boldsymbol\phi}$ that produces the four outputs,
\begin{align}
&f_{\Theta}(x)=g_{\boldsymbol\phi}\!\left(z(\boldsymbol\theta,x)\right)=\\\nonumber
&\big(\Re e \mathcal{H},\ \Re e \mathcal{E},\ \Re e \mathcal{\widetilde{H}},\ \mathrm{DVCS}\big),~\Theta\equiv(\boldsymbol\theta,\boldsymbol\phi).
\end{align}

Gradients of expectation values with respect to circuit parameters are evaluated by the parameter‑shift rule~\cite{mitarai2018quantum},
\begin{equation}
\frac{\partial}{\partial\theta_k}\,z_j(\boldsymbol\theta,x)
=\tfrac{1}{2}\!\left[z_j(\boldsymbol\theta_{k}^{+},x)-z_j(\boldsymbol\theta_{k}^{-},x)\right],
\label{eq:psr}
\end{equation}
where
\begin{equation}
\boldsymbol\theta_{k}^{\pm}:\ \theta_k\mapsto \theta_k\pm \tfrac{\pi}{2},\nonumber
\end{equation}
enabling standard first‑order optimizers. The QDNN is trained with the same physics‑informed objective as the CDNN: for measured points $\{(x_i,\phi_i,F_i^{\rm data},\sigma_i)\}_{i=1}^N$,
\begin{equation}
\mathcal{L}(\Theta)=\frac{1}{N}\sum_{i=1}^N
\frac{\Big[F^{\rm pred}\!\left(\phi_i; f_{\Theta}(x_i)\right)-F_i^{\rm data}\Big]^2}{\sigma_i^2},
\label{eq:loss}
\end{equation}
where $F^{\rm pred}$ is computed from the BKM10 twist‑2 DVCS/BH decomposition with the DVCS term taken $\phi$‑independent at this order.

For a fixed forward model $F_{\mathrm{pred}}$, minimization of the objective defined in Eq.~(\ref{eq:loss}) is mathematically equivalent to a weighted least-$\chi^2$ fit under the assumption of Gaussian experimental uncertainties. In the local extraction performed here, the network outputs $f_{\Theta}(x_B,Q^2,t)$ play the role of bin-wise fit parameters representing effective twist-2 Compton Form Factors (CFFs), and the CDNN or QDNN architecture serves as a parameterization together with a gradient-based optimizer.

The purpose of the DNN framework is not to replace least-$\chi^2$ fitting, but to provide an improved parameterization and optimization strategy for the same statistical objective. By reparameterizing the physical fit parameters as the output of a differentiable model and minimizing $\chi^2$ in the model-parameter space, the approach (i) enables clear uncertainty propagation through the forward model, (ii) extends naturally to higher-dimensional extractions involving additional CFF components or observables without restructuring the minimization procedure, and (iii) permits controlled inductive bias and regularization through architectural design, which can improve stability in sparse or high-uncertainty kinematic regimes. For context on harmonic-analysis based extractions of twist-2 CFFs, see Ref.~\cite{Shiells:2022HarmonicCFF}.

Encoding classical inputs into an $n$-qubit state space (Hilbert-space dimension $2^n$) and applying
entangling gates defines a nonlinear feature map that can be difficult for shallow classical models
to emulate for certain data distributions~\cite{biamonte2017quantum,schuld2015introduction,havlicek2019supervised,schuld2018quantum}.
Trainability of \emph{parameterized (variational) quantum circuits} (PQCs) is task- and ansatz-dependent:
while highly expressive PQCs exist~\cite{du2020expressive} and quantum-enhanced feature spaces have been
constructed~\cite{havlicek2019supervised}, increasing circuit depth can lead to vanishing gradients
(\emph{barren plateaus})~\cite{mcclean2018barren}.
To balance expressivity and optimization stability, we use depth-controlled \emph{Strongly Entangling Layer}
(SEL) blocks together with layer-wise depth growth and depth-scaled parameter initialization to stabilize
gradients~\cite{abbas2021power,cerezo2021cost}.
All comparisons in this work are performed on a \emph{quantum simulator}; reported floating-point operation
(FLOP) counts therefore refer to the classical cost of simulated state-vector evolution, whereas on
quantum hardware the relevant resources are circuit depth, the number and connectivity of two-qubit gates,
and the number of measurement shots.
Finally, broader results in \emph{quantum machine learning} (QML)—including quantum-enhanced feature maps and
generative models~\cite{havlicek2019supervised,zoufal2019quantum,biamonte2017quantum}—motivate this approach,
but do not imply an advantage for every regression task.

While simulators are ideal for prototyping and benchmarking, their classical cost scales
exponentially with system size. In a brute-force state-vector simulation, an $n$-qubit pure state
requires storing $2^n$ complex amplitudes, i.e., $\mathcal{O}(2^n)$ memory; with double-precision
complex numbers this corresponds to $\sim 16\times 2^n$ bytes (about $16\,\mathrm{PB}$ at $n=50$),
doubling with each additional qubit~\cite{Haner2017,Wang2021Sunway}. Consequently, exact simulation
of generic, highly entangling circuits becomes impractical beyond $\mathcal{O}(50)$ qubits except
on the largest high-performance computing (HPC) platforms~\cite{DeRaedt2025JUQCS50}, whereas
structure-exploiting methods (e.g., tensor-network simulators) can extend classical reach only in
regimes where the circuit generates limited entanglement~\cite{Seitz2023TTN}.
Beyond this scale---for example, at larger qubit counts and depths, or when assessing
beyond-classical performance---real \emph{noisy intermediate-scale quantum} (NISQ) hardware becomes
essential~\cite{Preskill2018NISQ}. Moreover, only hardware experiments expose realistic noise and
sampling effects and enable the development and validation of quantum error-mitigation and hybrid
workflows~\cite{Cai2023QEM}. Since the present study is focused on the algorithmic comparison of
model classes for DVCS CFF extraction, we therefore restrict all QDNN results to simulator-based
experiments.

Section~\ref{sec:CFFExtraction} benchmarks a \emph{Basic} QDNN (isocapacity with the CDNN) and a \emph{Full} QDNN that introduces quantum‑specific optimizations (entanglement range control, layer‑wise depth growth, depth‑scaled initialization). Section~\ref{sec:dvcs-qq} defines a DVCS quantum qualifier that predicts when the QDNN should outperform the CDNN, and Sec.~\ref{sec:cff-extract-exp-data} applies the qualifier‑guided extraction to experimental data.

\section{\label{sec:ExperimentalData}Experimental Data and Pseudodata Generation}

\begin{table*}[htb!]
\caption{\label{tab:data_summary}Summary of the DVCS data from JLAB used in this analysis. The table includes information on the data source and the kinematic range covered by the helicity-independent cross section.}
\begin{ruledtabular}
\begin{tabular}{lcccccc}
\multicolumn{1}{c}{Experiment} & Publication Year & $E_{beam}$ (GeV) & $Q^2$ (GeV$^2$) & $-t$ (GeV$^2$) & $x_B$ & Number of Points\\
\hline
Hall A E12-06-114 \cite{Georges2022} & 2022 & 4.487 - 10.992 & 2.71 - 8.51 & 0.204 - 1.373 & 0.363 - 0.617 & 1080\\
Hall A E07-007 \cite{Defurne2017} & 2017 & 3.355 - 5.55 & 1.49 - 2. & 0.177 - 0.363 & 0.356 - 0.361 & 404 \\
Hall A E00-110 \cite{Defurne2015} & 2015 & 5.75 & 1.82 - 2.37 & 0.171 - 0.372 & 0.336 - 0.401 & 468\\
Hall B e1-DVCS1 \cite{Jo2015} & 2015 & 5.75 & 1.11 - 3.77 & 0.11 - 0.45 & 0.126 - 0.475 & 1933\\ 
\end{tabular}
\end{ruledtabular}
\end{table*}

The CFF extraction analyses detailed in the remainder of this paper utilize DVCS experimental data from Hall A~\cite{Defurne2015, Defurne2017, Georges2022} and Hall B~\cite{Jo2015} at Jefferson Lab (JLab). The dataset, summarized in Table~\ref{tab:data_summary}, includes both helicity-independent and helicity-dependent cross sections. However, our focus is exclusively on the helicity-independent (unpolarized) cross sections for more direct comparison to previous work \cite{Diaz:2025}. The data is finely binned in a fourfold differential cross section parameterized by \( Q^2 \), \( x_B \), \( t \), and \( \phi \), comprising a total of 3,885 data points for analysis. 

The dataset spans measurements from both the 6 GeV and 12 GeV JLab experimental programs. The 12 GeV data extend to higher values of \( Q^2 \), reaching up to 8.4 GeV\(^2\), while the 6 GeV-era data cover lower ranges. The majority of the data points originate from the e1-DVCS1 experiment at Hall B~\cite{Jo2015}, which operated with a fixed beam energy of 5.75 GeV and probed a \( Q^2 \) range from 1.0 to 4.6 GeV\(^2\). These experimental data sets serve as a crucial foundation, not only for generating realistic pseudodata, but also to establish initial benchmarks for developing and validating our machine learning techniques, enabling systematic comparisons between CDNNs and QDNNs under controlled conditions.

We generate numerous pseudodata sets based directly on the experimental data sets. These pseudodata sets allow us to precisely control the ``true'' CFF values, offering a controlled environment for closure testing and evaluating the accuracy and performance of our neural network models before their application to real experimental data, where the true values are unknown. Our pseudodata sets are generated using the following \textit{basic} model function:
\begin{equation} G(x_B,t) = (a x_B^2 + bx_B) e^{ct^2+ dt + e} + f, \label{equ:model-par} \end{equation}
where \( G(x_B,t) \) represents the CFFs and DVCS cross section components: \( \Re e \mathcal{H}, \Re e \mathcal{E}, \Re e \widetilde{\mathcal{H}}, \) and \( DVCS \). The parameters \( \{a, b, c, d, e, f\} \) for each case are listed in Table~\ref{tab:table_par_cff}, and are based on initial fits to experimental data. The structure of this model function allows for linear or non-linear changes in $x_B$ with exponential flexibility in $t$.  The parameterization has no physical meaning as the model is purely designed for testing the DNN extractions independently of any theoretical model, so as to reduce the likelihood of bias in the testing phase.

\begin{table}[tb!]
\caption{\label{tab:table_par_cff}%
The parameters used in Eq.~\eqref{equ:model-par} to generate the pseudodata set. 
}
\begin{ruledtabular}
\begin{tabular}{ccccccc}
CFFs & $a$ & $b$ & $c$ & $d$ & $e$ & $f$ \\
\hline
$\Re e \mathcal{H}$ & -4.41 & 1.68 & -9.14 & -3.57 & 1.54 & -1.37 \\
$\Re e \mathcal{E}$ &  144.56 & 149.99 & 0.32 & -1.09 & -148.49 & -0.31   \\
$\Re e \widetilde{\mathcal{H}}$ & -1.86  & 1.50 & -0.29 & -1.33 & 0.46 & -0.98  \\
$DVCS$ & 0.50 & -0.41 & 0.05 & -0.25 & 0.55 & 0.166 \\
\end{tabular}
\end{ruledtabular}
\end{table}

To compute the photon leptoproduction cross section described in Section~\ref{sec:TheoryFramework}, the CFFs are taken from the \textit{basic} model. Within each kinematic bin, the cross section is randomly sampled from a Gaussian distribution centered on the calculated value at each \( \phi \) point, with the standard deviation corresponding to experimental uncertainties. The generated kinematic settings and \( \phi \) points match those of the experimental data used in this study, ensuring that acceptance effects and statistical limitations are properly accounted for. 

The uncertainties introduced in the pseudodata are taken from the reported experimental errors from JLab data~\cite{Georges2022, Defurne2015, Defurne2017, Jo2015}, which include both statistical and systematic contributions. As a result, the generated pseudodata closely replicate real-world experimental conditions that influence the measured cross sections.

To account for statistical fluctuations in the data analysis, a set of replica datasets is produced for each pseudodata configuration. In each replica, the cross section at the central value of each bin is resampled using Gaussian noise consistent with the original experimental uncertainty. This ensemble approach enables a robust estimation of the variability in extracted observables due to measurement uncertainty.

With this ensemble of realistic pseudodata in hand, we proceed to the central task of our study: extracting CFFs using both classical and quantum deep neural networks, and directly comparing their respective performance across the full range of kinematics.

\section{\label{sec:CFFExtraction}Pseudodata Comparison}

In this section, we focus exclusively on local fitting and compare our two deep neural network approaches---CDNNs and QDNNs---for extracting CFFs from the pseudodata.

Traditional local fits extract CFFs independently at each kinematic point, typically using data binned in the azimuthal angle \( \phi \), which describes the relative orientation of the lepton and hadron scattering planes. The fitting procedure is based on the helicity amplitudes and is performed separately for each kinematic bin. While this method avoids assumptions about underlying functional forms, it can lead to non-uniqueness in extracted CFFs due to limited constraints, often resulting in large systematic uncertainties. DNN-based fitting methods offer an alternative approach by using statistical correlations across the dataset to improve extraction precision. Unlike analytical parameterizations that impose model-specific constraints, these neural network methods learn directly from experimental data without requiring predefined functional dependencies or initial guesses of the CFFs.

We compare the performance of CDNNs and QDNNs in extracting CFFs, assessing their effectiveness through a testing phase using the pseudodata generated previously described. The main evaluation metrics used are accuracy and precision. Accuracy \( \epsilon(k,Q^2,x_B,t) \) quantifies how closely the mean of the DNN-extracted CFFs aligns with the true CFF values used in pseudodata generation:
\begin{equation}
    \epsilon(k,Q^2,x_B,t) = |CFF_{\text{DNN}} - CFF_{\text{true}}|.
    \label{eq:accuracy}
\end{equation}
This metric is actually the absolute value of the residual used to define accuracy, which remains meaningful even if \( CFF_{\text{DNN}} \) and \( CFF_{\text{true}} \) are not less than 1. The precision \( \sigma(k,Q^2,x_B,t) \) reflects the variation in extracted CFFs across multiple replicas:
\begin{equation}
    \sigma (k,Q^2,x_B,t) =  \sqrt{\frac{\sum_{i} \left(CFF^i_{\text{DNN}}-\overline{CFF}_{\text{DNN}}\right)^2}{N}}.
    \label{eq:precision}
\end{equation}
The generating function used to produce the \textit{true} CFF values follows a \textit{basic} model from Eq.~\eqref{equ:model-par} using parameters from Table \ref{tab:table_par_cff}, and the pseudodata we extract the CFFs from is described in Sec. \ref{sec:ExperimentalData}.

In the next subsections, we rigorously compare CDNN and QDNN approaches, evaluating their ability to extract CFFs with through accuracy, precision, and stability.  Further, we consider two QDNN models: the Basic QDNN described in the next section (Sec. \ref{subsec:bqdnn}) and the Full QDNN described in following section (Sec. \ref{subsec:leveraging-quantum}). The Basic QDNN is designed to mirror the CDNN model in architecture and complexity, serving as a baseline comparison without quantum-specific optimization. The Full QDNN is designed to fully take advantage of the quantum potential by carefully mitigating quantum-specific challenges and incorporating quantum-specific optimization schemes. The details of the architecture and construction of the CDNN and QDNNs, the method of extraction, and the corresponding results are presented in the following subsections.

\subsection{Basic QDNN}
\label{subsec:bqdnn}

We begin by performing the CFF extraction using a CDNN model and a Basic QDNN model (which we refer to as just QDNN in this section). This extraction serves to gauge the immediate advantages and disadvantages of a direct transition to a quantum-based model before considering quantum-specific effects. 

All the DNN models used for extraction in this paper have a three-dimensional input ($x_B$, $Q^2$, $t$), the kinematic values of a given kinematic bin, and a four-dimensional output ($\Re e\mathcal{H}$, $\Re e\mathcal{E}$, $\Re e\widetilde{\mathcal{H}}$, $DVCS$), the four CFFs. The methodology of the local CFF fitting used here follows a standard maximum-likelihood estimation approach except with the added benefit of leveraging neural networks optimization algorithms and flexibility.  Such fits were first applied to local CFF extraction by Kumeri\v{c}ki et al.~\cite{KMS11}.  For more detail on modern methodology and error propagation see \cite{Keller:2025eup}. The DNNs use a custom, physics-informed loss function that predicts the cross section $F$ using the outputted three CFFs and DVCS term with the BKM10 formalism (Sec. \ref{sec:TheoryFramework}) and computes the mean squared error of the predicted cross section against the true cross section.

\begin{figure}[ht!]
    \centering
        \includegraphics[width=0.48\textwidth]{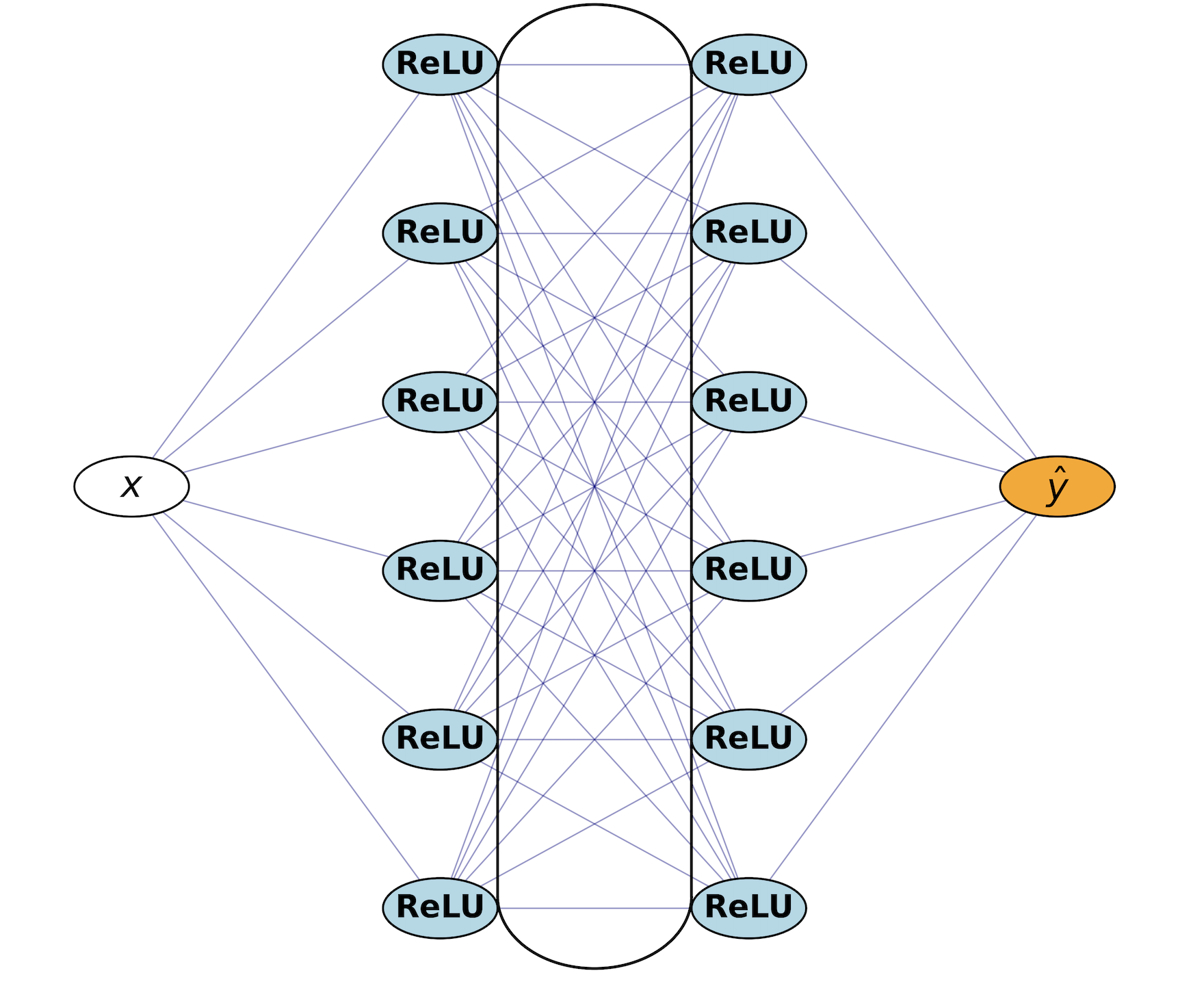}
        \caption{Schematic of the CDNN used for local CFF regression. The network takes a
three-dimensional kinematic input $(x_B,Q^2,t)$ and maps it through an input layer,
eight fully connected hidden layers (64 neurons each) with ReLU activations, to a
four-dimensional output $(\Re e\,\mathcal{H},\Re e\,\mathcal{E},\Re e\,\widetilde{\mathcal{H}},\mathrm{DVCS})$. 
Only the first and last hidden layers are shown explicitly; intermediate layers are omitted for clarity.}
    \label{fig:CDNN-reg-arch}
\end{figure}
\begin{figure}[ht!]
    \centering
    \includegraphics[width=0.49\textwidth]{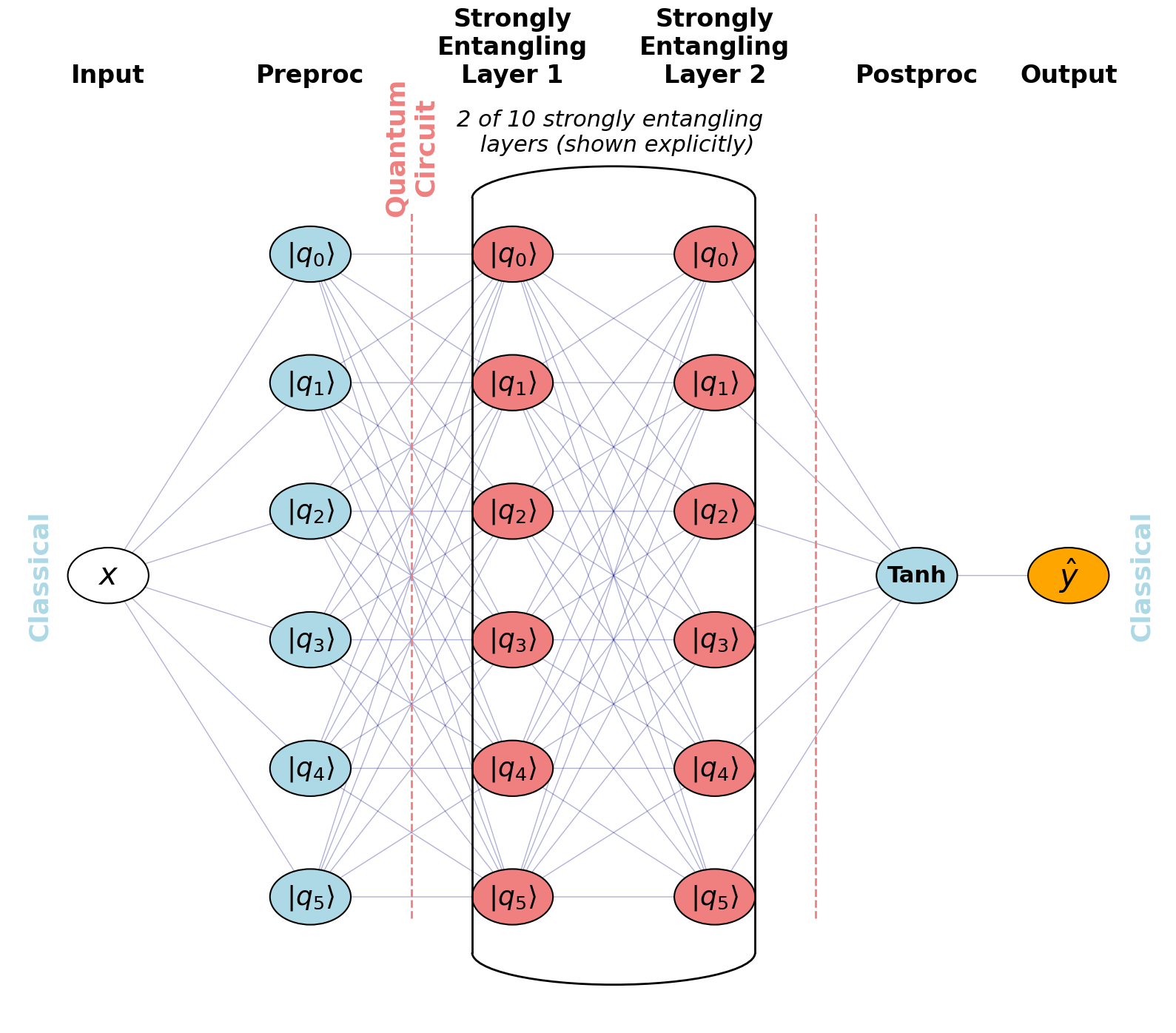}
    \caption{Schematic of the QDNN used for local CFF regression. A classical preprocessing
layer maps $(x_B,Q^2,t)$ to $n=6$ qubits via angle embedding. The circuit applies $L=8$
Strongly Entangling Layers (SELs); only a subset is drawn for clarity. Pauli-$Z$
expectation values are measured on each qubit to form a 6-dimensional classical vector,
which is passed to a small classical head to produce four outputs
$(\Re e\,\mathcal{H},\Re e\,\mathcal{E},\Re e\,\widetilde{\mathcal{H}},\mathrm{DVCS})$.}
    \label{fig:QDNN-reg-arch}
\end{figure}
The CDNN model used for the extraction of the CFFs is based closely on the CDNN regression feedforward architecture shown in Fig.~\ref{fig:CDNN-reg-arch}. The input layer of the CDNN maps the three-dimensional input layer ($x_B$, $Q^2$, $t$) to 64 neurons. Then, there are 8 hidden layers, each with 64 neurons with rectified linear unit (ReLU) activation ($\mathrm{ReLU}(z)=\max(0,z)$). Finally, the output layer maps the 64 features to a four-dimensional output ($\Re e\mathcal{H}$, $\Re e\mathcal{E}$, $\Re e\widetilde{\mathcal{H}}$, $DVCS$). The QDNN is constructed based on the architecture of the CDNN and is similar to the architecture shown in Fig.~\ref{fig:QDNN-reg-arch}. The input layer of the QDNN is a classical linear preprocessing layer, mapping the three-dimensional input to 6 qubits, which is equivalent to the 64 neurons used in the CDNN (Sec. \ref{sec:QuantumDNNs}). The input data is then embedded into quantum state space using angle embedding, which encodes the features as rotation angles on the qubits. Next, there are 8 hidden Strongly Entangling Layers, consisting of rotations and entanglers. Finally, the expectation value of the Pauli-Z operator for each qubit is measured. In the classical postprocessing layers, the 6 quantum outputs are mapped to 64 neurons, followed by a Tanh activation, then another linear layer to the final four-dimensional output.

To enable a controlled benchmark, we keep the forward model, loss, training protocol, and input/output structure fixed, and choose architectures with comparable forward-pass FLOP estimates. The QDNN uses substantially fewer trainable parameters, so performance differences reflect model class and inductive bias rather than simply parameter count (Sec. \ref{subsubsec:model-complex}). Subsequently, we apply both models to extract Compton form factors (CFFs) from pseudodata and present a comprehensive error analysis of the results (Sec. \ref{subsubsec:cff-extraction-results-bqdnn}).

\begin{table*}[ht!]
    \centering
    \caption{\label{tab:model-complex-compar} Comparison of model complexity between the CDNN and the QDNN. The table lists key architectural metrics, including depth (layer composition), width (neurons or qubits), number of trainable parameters, and estimated floating-point operations (FLOPs). The QDNN includes both quantum and classical layers, with a 64-dimensional quantum encoding based on 6 qubits.}
    \begin{ruledtabular}
    \begin{tabular}{ccc}
        Metric & CDNN & QDNN \\
        \hline
        Depth & 10 layers (2 in/out + 8 hidden) & 11 layers (8 quantum + 3 classical) \\
        Width & 64 neurons & 6 qubits (64-dimensional) \\
        Trainable Parameters & 33796 parameters & 876 parameters \\
        FLOPs & 67008 FLOPs & 74688 FLOPs
    \end{tabular}
    \end{ruledtabular}
\end{table*}

\subsubsection{Model Complexity}
\label{subsubsec:model-complex}

A fair comparison between the CDNN and QDNN requires ensuring that neither architecture gains an artificial advantage from simply being larger or more computationally powerful. We therefore quantify and compare two standard measures of model complexity---the number of trainable parameters and the approximate floating-point operation count (FLOPs) for a forward pass. These metrics characterize a model's representational capacity and computational cost, respectively, and provide a baseline for assessing whether the CDNN and Basic QDNN operate at comparable levels of complexity despite belonging to fundamentally different paradigms.

Direct comparisons between classical and quantum models are inherently subtle. A QDNN may have far fewer tunable parameters, yet its entangling structure can encode highly non-linear transformations that would require a significantly larger classical network to approximate. Likewise, FLOPs are not strictly well-defined for quantum circuits, which evolve qubit states via unitary operations rather than classical arithmetic. Nevertheless, computing these metrics offers a useful reference point and ensures that neither architecture is advantaged by a disproportionate model size.

\paragraph*{Trainable Parameters:} For a fully connected layer with input dimension $M$ and output dimension $N$, the number of trainable parameters is
\begin{equation}
    P = MN + N.
\end{equation}

The CDNN consists of an input layer ($3\!\to\!64$), eight hidden layers ($64\!\to\!64$), and an output layer ($64\!\to\!4$). The total number of trainable parameters is
\begin{equation}
    \begin{split}
        P_{\text{CDNN}} &= (3\cdot 64 + 64)
        + 8(64\cdot 64 + 64)
        + (64\cdot 4 + 4) \\
        &= 33796.
    \end{split}
\end{equation}

The QDNN includes a preprocessing layer ($3\!\to\!6$), eight Strongly Entangling Layers (SELs), and postprocessing layers ($6\!\to\!64\!\to\!4$). Each SEL applies three parameterized rotations per qubit:
\begin{equation}
    P_{\text{SEL}} = 8 \cdot 6 \cdot 3 = 144.
\end{equation}
The angle-embedding step contains no trainable parameters. The total parameter count is
\begin{equation}
    \begin{split}
        P_{\text{QDNN}} &= (3\cdot 6 + 6) + 144
        + (6\cdot 64 + 64)
        + (64\cdot 4 + 4) \\
        &= 876.
    \end{split}
\end{equation}

As expected, the QDNN uses far fewer parameters due to its ability to encode a $2^n$-dimensional Hilbert space in $n$ qubits.

\paragraph*{Floating-Point Operation Count:}  
For a classical dense layer, the forward-pass FLOP count is
\begin{equation}
    C = 2MN.
\end{equation}

Including ReLU activations (64 FLOPs each), the total FLOPs are
\begin{equation}
    \begin{split}
        C_{\text{CDNN}} &= (2\cdot 3\cdot 64 + 64)
        + 8(2\cdot 64\cdot 64 + 64)
        + (2\cdot 64\cdot 4) \\
        &= 67008.
    \end{split}
\end{equation}

Quantum layers act on the $2^6=64$-dimensional state vector. Approximating FLOPs for preprocessing, eight SELs, postprocessing layers, and a Tanh activation (64 FLOPs), we obtain
\begin{equation}
    \begin{split}
        C_{\text{QDNN}} &= (2\cdot 3\cdot 64)
        + 8(2\cdot 64\cdot 64) \\
        &\mathrel{\phantom{=}}+ \bigl(2\cdot 64\cdot 64 + 64 + 2\cdot 64\cdot 4\bigr) \\
        &= 74688.
    \end{split}
\end{equation}

The resulting FLOP estimate is of the same order as the CDNN.

Table~\ref{tab:model-complex-compar} summarizes the comparison. Despite their structural differences, the CDNN and QDNN exhibit similar effective complexity when assessed through parameter counts and approximate FLOPs, ensuring that our performance comparison is not biased by model size or computational load.

\begin{figure}[ht!]
    \centering
    \includegraphics[width=0.45\textwidth]{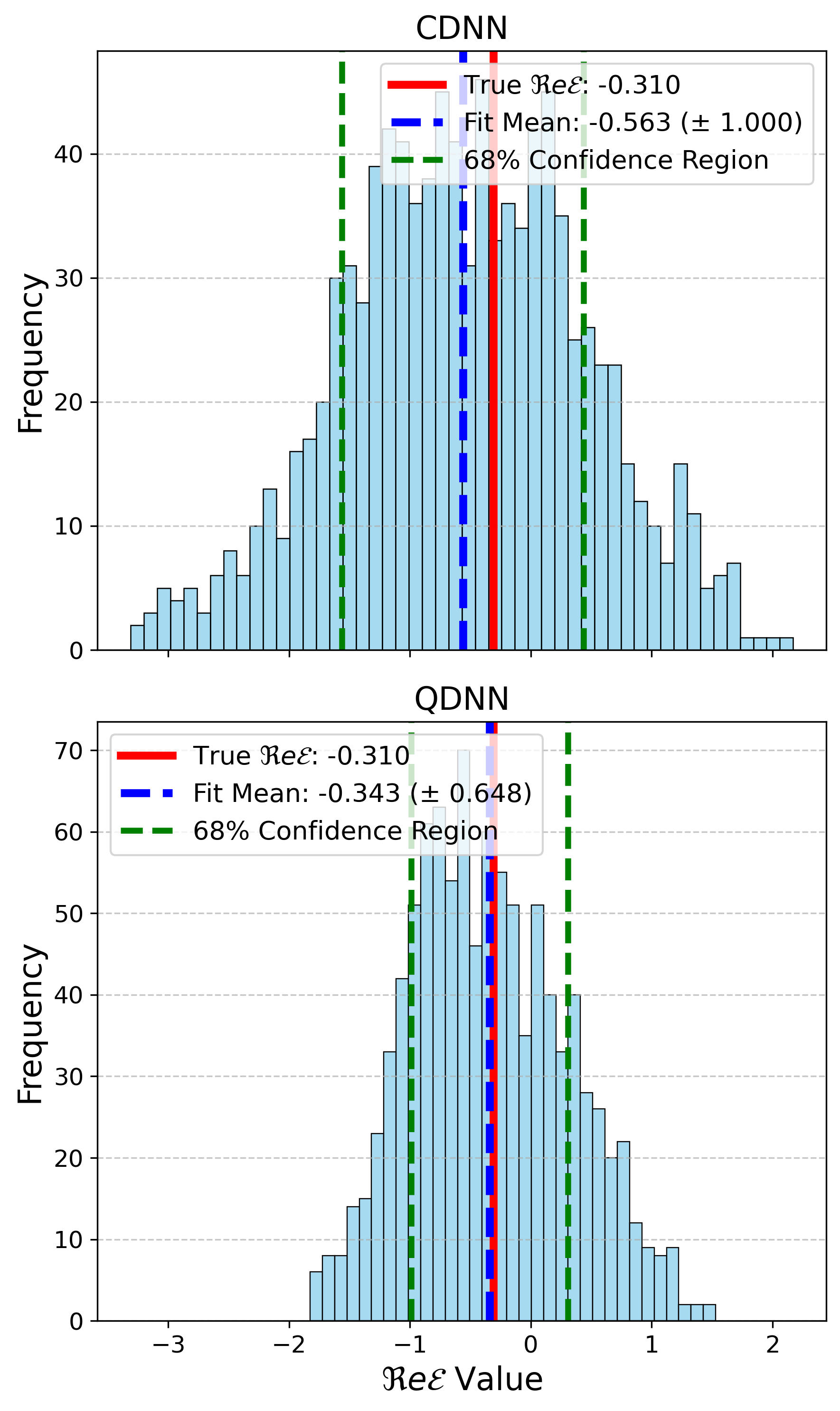}
    \caption{Distributions of the extracted $\Re e\mathcal{E}$ from noisy replicas of cross section pseudodata generated from kinematics $k=5.75$ GeV, $Q^2 = 2.22$ GeV$^2$, $x_B = 0.333$, $t = -0.16$ GeV$^2$ (Set 144). The histograms display improved accuracy and precision by the QDNN.}
    \label{fig:ReE-dist}
\end{figure}

\begin{figure*}[htp!]
    \centering
    \subfloat[]{\includegraphics[width=0.45\textwidth]{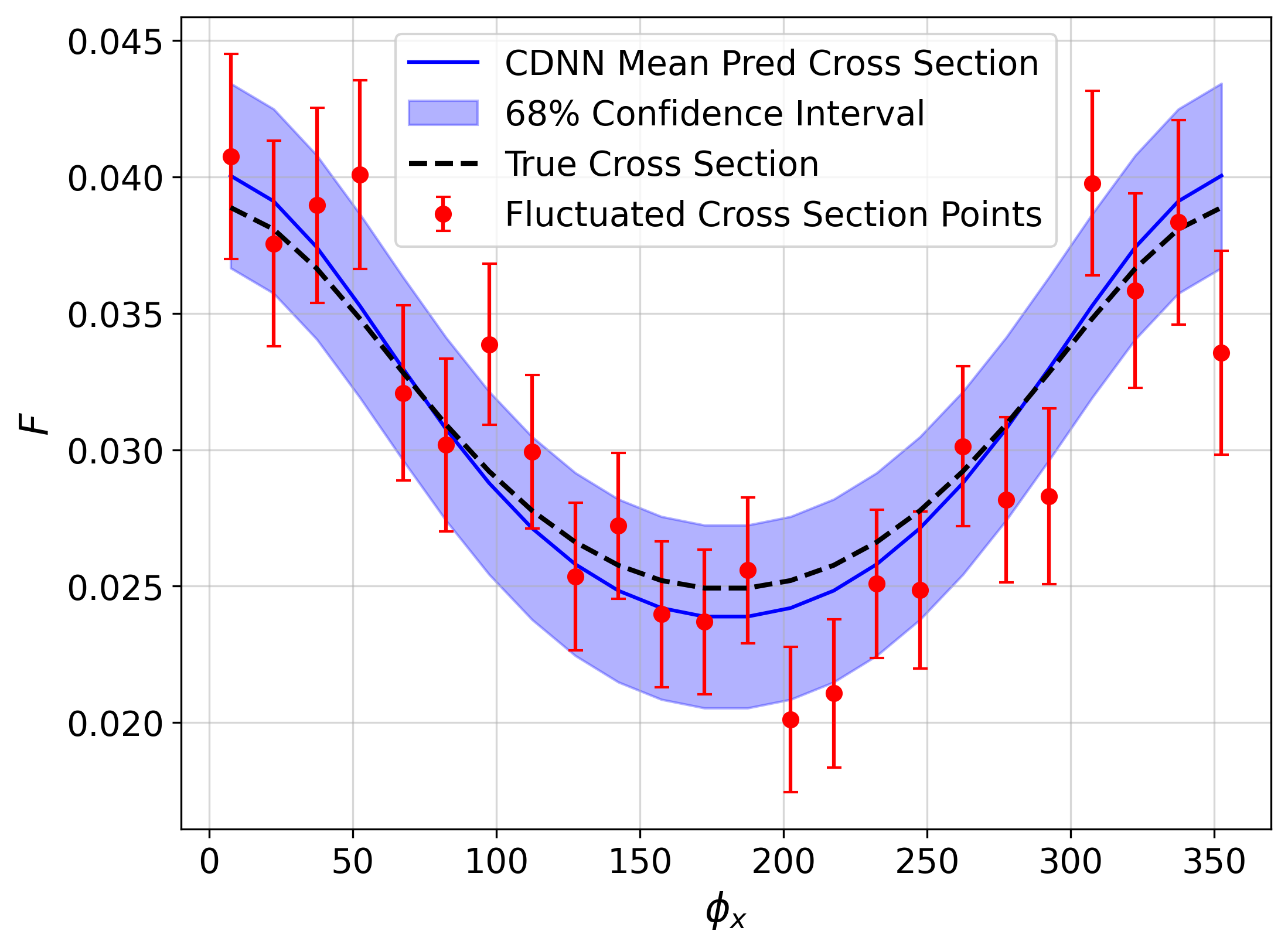}\label{fig:c-fvsphi-1}}
    \hfill
    \subfloat[]{\includegraphics[width=0.45\textwidth]{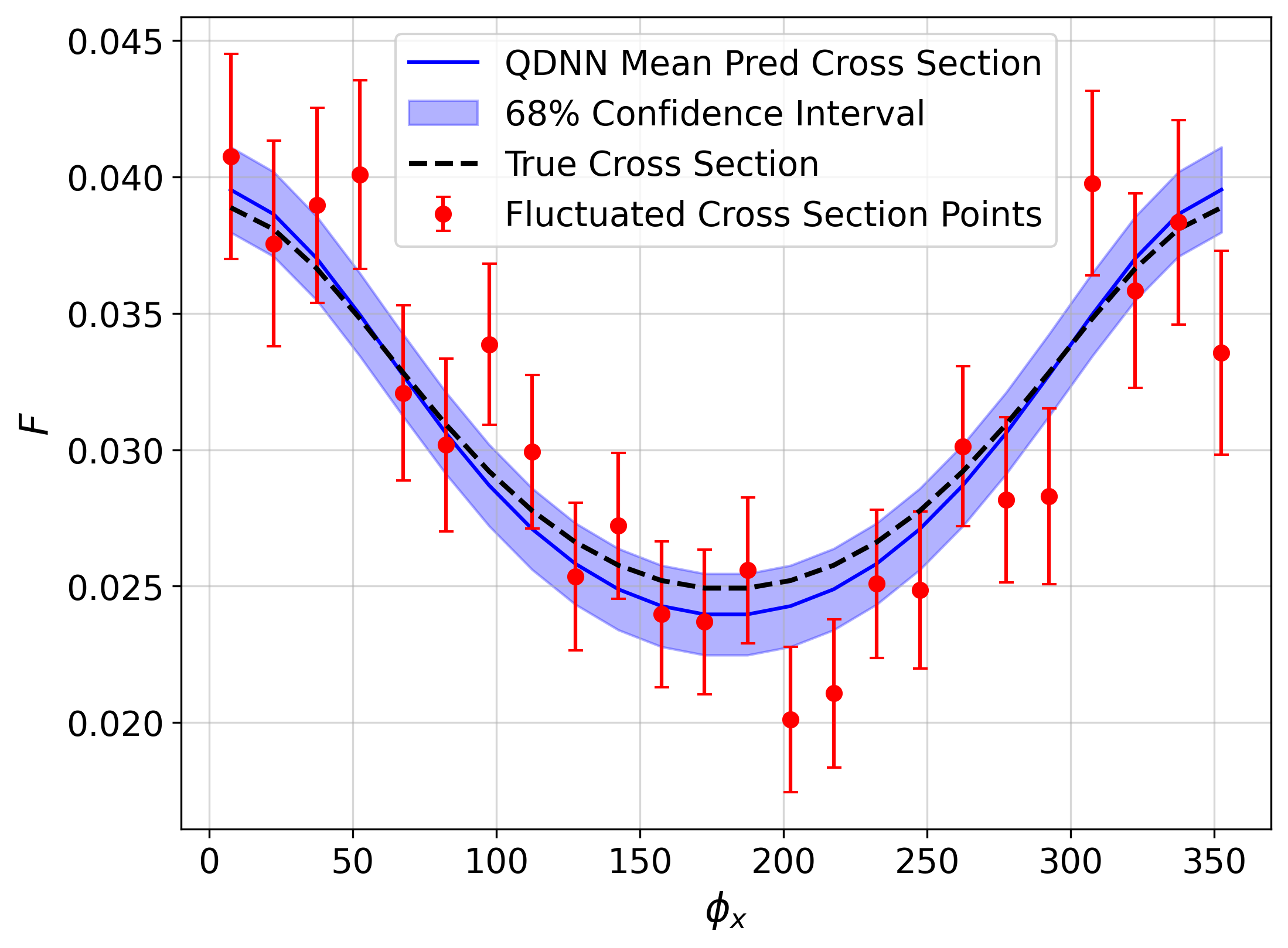}\label{fig:q-fvsphi-1}}
    \caption{Predicted cross sections from the (a) CDNN and (b) QDNN fits of noisy replicas of cross section pseudodata (red points) generated from kinematics $k=8.521$ GeV, $Q^2 = 3.65$ GeV$^2$, $x_B = 0.367$, $t = -0.20459$ GeV$^2$ (Set 26). }
    \label{fig:fvsphi-1}
\end{figure*}
\begin{figure*}[htp!]
    \centering
    \subfloat[]{\includegraphics[width=0.45\textwidth]{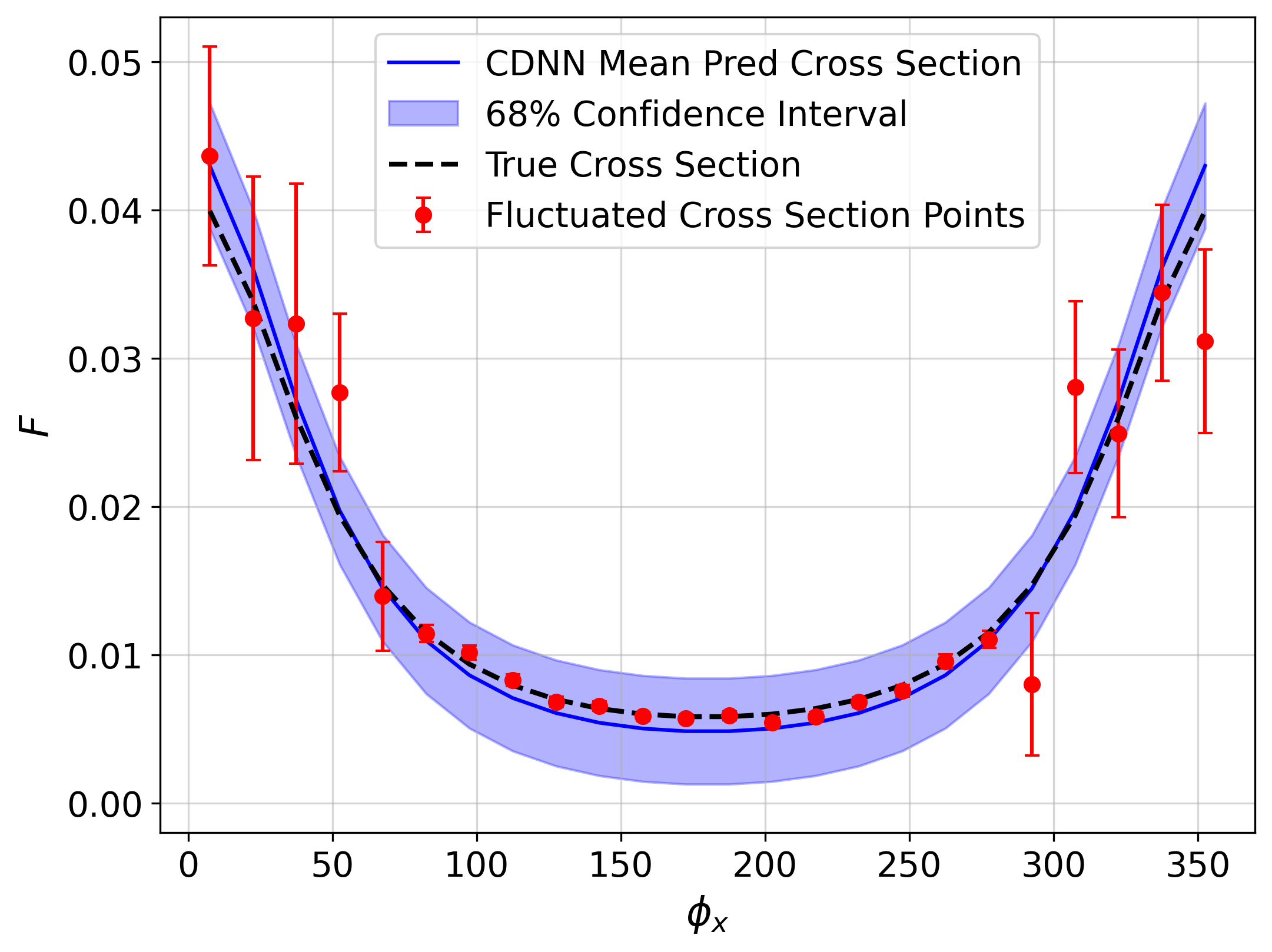}\label{fig:c-fvsphi-3}}
    \hfill
    \subfloat[]{\includegraphics[width=0.45\textwidth]{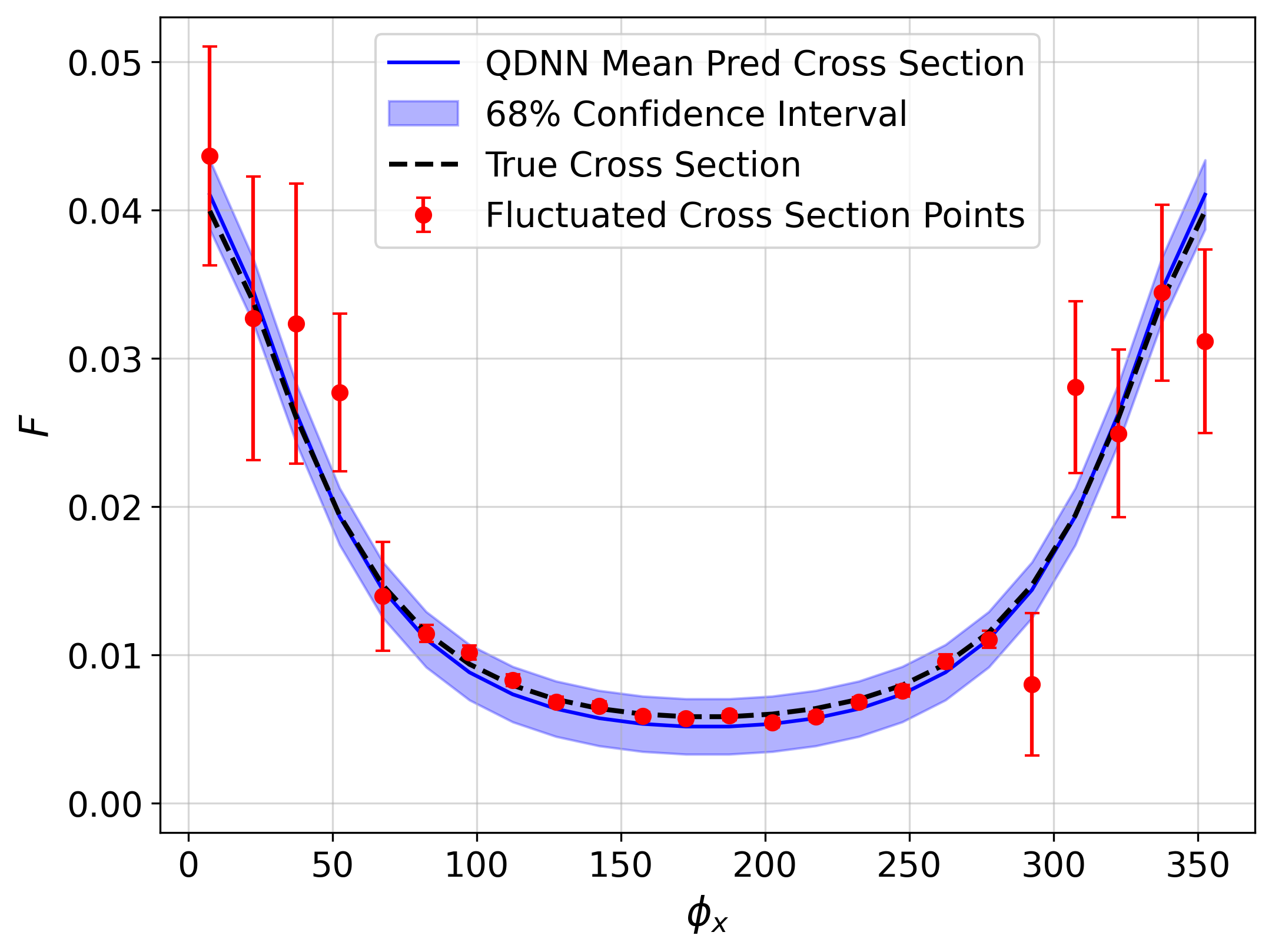}\label{fig:q-fvsphi-3}}
    \caption{Predicted cross sections from the (a) CDNN and (b) QDNN fits of noisy replicas of cross section pseudodata (red points) generated from kinematics $k=5.75$ GeV, $Q^2 = 2.48$ GeV$^2$, $x_B = 0.399$, $t = -0.45$ GeV$^2$ (Set 165). In both cases there is a small improvement in proximity to the true with a clear narrowing of the error band for the QDNN.}
    \label{fig:fvsphi-3}
\end{figure*}

\begin{figure}
    \centering
    \includegraphics[width=0.45\textwidth]{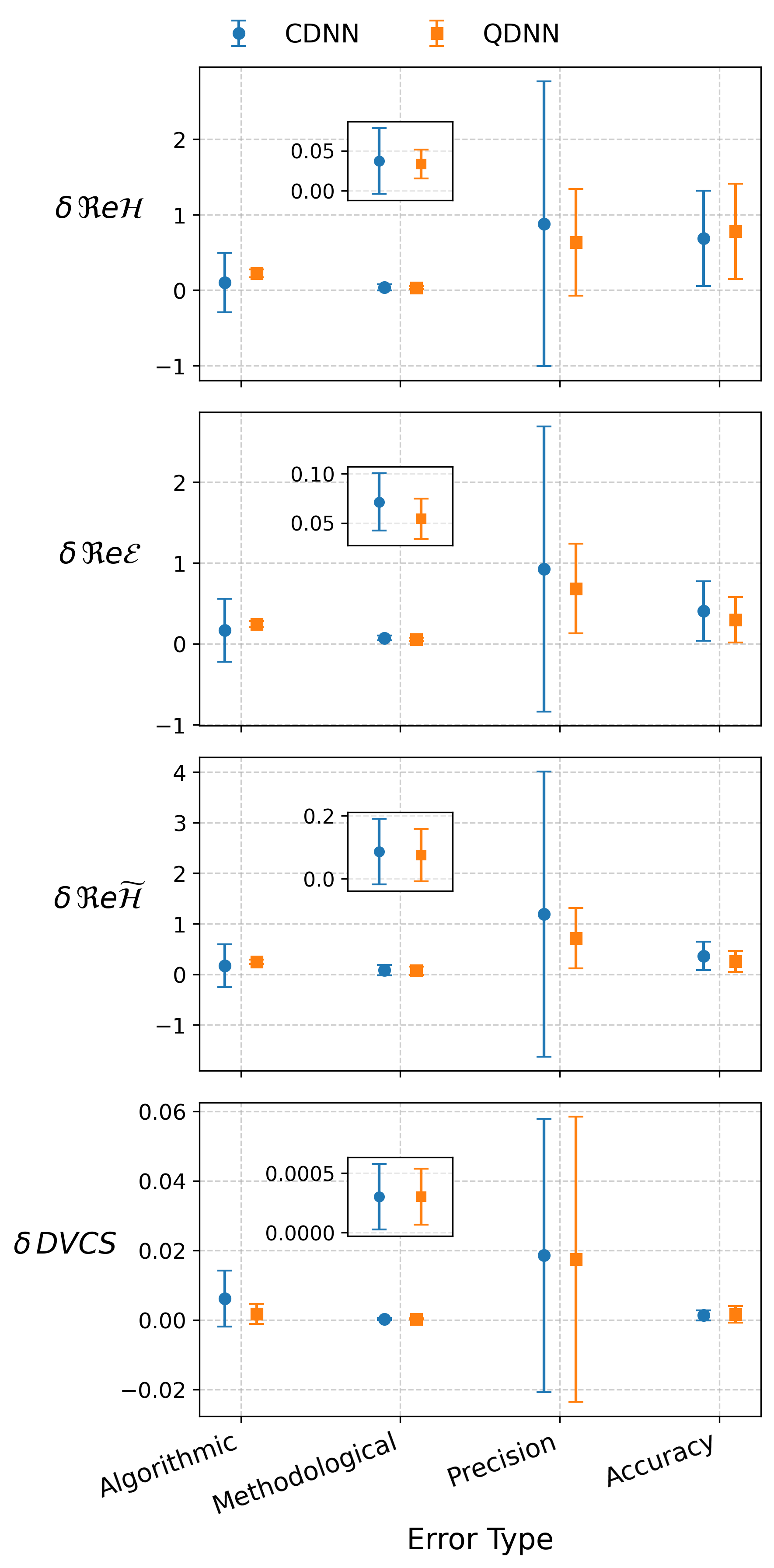}
    \caption{Comparative error distributions of the CDNN (blue) and QDNN (orange) extraction across all kinematic bins. The points and error bars show histogram mean and error. Algorithmic error isolates uncertainty in the extraction process itself. The QDNN shows higher algorithmic error due to quantum-specific challenges, addressed in Sec. \ref{subsec:leveraging-quantum}. Methodological error captures uncertainty arising from model assumptions. The QDNN consistently demonstrates lower methodological error, highlighting its robustness to systematic uncertainties. Precision reflects consistency across runs, regardless of proximity to the true value. The QDNN achieves higher precision for all CFFs, ensuring more stable and reliable extractions even when accuracy is lower. Accuracy measures how close the DNN extractions are to the true CFFs. The QDNN shows improved accuracy for $\Re e\mathcal{E}$ and $\Re e\widetilde{\mathcal{H}}$.}
    \label{fig:CFF_error_results}
\end{figure}

\subsubsection{CFF Extraction Test}
\label{subsubsec:cff-extraction-results-bqdnn}

Having established the architectures and complexity of the CDNN and Basic QDNN models, we now present the results of their performance in extracting CFFs from the pseudodata sets. Figure~\ref{fig:ReE-dist} shows representative histograms of the $\Re e\mathcal{E}$ extraction from 1000 noisy replicas of the cross-section pseudodata for a fixed kinematic setting. In this case, the QDNN exhibits improvement in both accuracy (proximity to the true) and precision (width of the distribution).

To quantify overall performance, we define a reduced $\chi^2$ metric, $M_{\chi^2}$, as the sum of the reduced $\chi^2$ values for each CFF across all 195 kinematic bins.
\begin{equation}
M_{\chi^2} \equiv \sum_{F \in \{\Re e\mathcal H,\Re e\mathcal E,\Re e\widetilde{\mathcal H},\mathrm{DVCS}\}}
\frac{1}{N_{\rm bins}} \sum_{b=1}^{N_{\rm bins}}
\frac{\big(\bar F_b - F^{\rm true}_b\big)^2}{\sigma^2(F_b)}.
\end{equation}
A lower $M_{\chi^2}$ indicates a better fit, with a value of 4 corresponding to a perfect match. Averaging over all bins, we find $\overline{M}_{\chi^2}^{\text{CDNN}} \approx 78.97$ and $\overline{M}_{\chi^2}^{\text{QDNN}} \approx 28.18$, indicating that the QDNN outperforms the CDNN by approximately 64.3\%, even without leveraging quantum-specific enhancements.

The predicted cross sections using the predicted CFFs from the CDNN and QDNN are also analyzed. Using the framework described in Section \ref{sec:TheoryFramework}, we compute the predicted cross sections using the extracted CFFs and compare them to the true cross section values used to generate the pseudodata. In Figs. \ref{fig:fvsphi-1} and \ref{fig:fvsphi-3}, we present results for two representative kinematic bins, plotting the mean predicted cross-section values as a function of the azimuthal angle $\phi$, along with their 68\% confidence intervals. The true cross-section curve and fluctuated true cross-section points with simulated experimental error from the pseudodata sets are overlaid for reference. A key observation from these figures is that the improved precision of the QDNN extraction leads to narrower confidence intervals in the predicted cross sections across all three bins due to the reduced uncertainty in the extracted CFFs. In these examples, there is also a small, observable improvement in the accuracy of the predicted cross section by the QDNN.

Understanding the sources of error is essential for robust model evaluation. A detailed analysis of the errors \cite{Keller:2025eup} in the CFF extractions is necessary to fully understand and quantify the differences between QDNN and CDNN approaches. We consider four distinct error metrics: algorithmic error, methodological error, precision, and accuracy. To compute the algorithmic error, we perform the extraction on 1000 \textit{identical} replicas of each bin and calculate the resulting spread of each CFF. Thus, algorithmic error represents error resulting from the limitations and inherent uncertainties of the extraction algorithm itself, rather than simulated experimental error in the data. To compute the methodological error, we introduce a spread in the parameters used to generate the CFFs (Table \ref{tab:table_par_cff}) and use the generating function [Eq.~\eqref{equ:model-par}] to generate CFFs sampled in a domain that approximately corresponds to the spread in the original extracted CFFs. Then, the methodological error is computed as the spread of the resulting residuals from the new extractions. Methodological error represents the uncertainty arising from the choice of model parameters and assumptions used in the extraction process. By introducing a spread in the parameters used to generate the CFFs and propagating this spread through the generating function, we assess how variations in these underlying assumptions influence the extracted CFF values. This approach allows us to estimate the systematic impact of modeling choices on the final extraction results, distinguishing it from purely statistical or algorithmic uncertainties. Finally, precision and accuracy are computed using Eqs. \eqref{eq:precision} and \eqref{eq:accuracy}, respectively. These metrics provide an evaluation of the performance of the DNNs on the original pseudodata set, with precision measuring the consistency of the extracted values across multiple replicas and accuracy reflecting the agreement between the extracted and true CFFs. For all four of these error metrics, a lower value indicates a lower error.

Figure~\ref{fig:CFF_error_results} presents a comparative analysis of algorithmic error, methodological error, precision, and accuracy for the DNN-based extraction of each CFF across all 195 kinematic bins. In the plot, the absolute mean deviation from the true value is shown for each contribution as a point, while the spread in that deviation is represented by the error bar for each point. A notable improvement in the three CFFs' precision is observed with the QDNN, whereas the other error contributions remain largely comparable between the two models. The QDNN shows an improved extraction accuracy for $\Re e\mathcal{E}$ and $\Re e\widetilde{\mathcal{H}}$ while the CDNN performs better for $\Re e\mathcal{H}$ and $DVCS$. Methodological errors are fairly small in both cases, but with a narrower distribution for the QDNN, indicating a bit better robustness to variations in the model parameters used to generate the CFFs.  The increase in precision from the QDNN is also evident in the predicted cross-section plots (Figs.~\ref{fig:fvsphi-1} and \ref{fig:fvsphi-3}) as visibly tighter confidence intervals. Overall, QDNN demonstrates improved total error performance compared to CDNN, underscoring its effectiveness in this application.

Although the QDNN outperforms the CDNN in terms of methodological error, precision, and accuracy, Fig.~\ref{fig:CFF_error_results} reveals that it exhibits a higher algorithmic error for the three CFFs. While QDNNs generally improve precision and accuracy, the simplified transformation of the CDNN into the basic QDNN introduced additional sources of uncertainty during training, leading to an overall increase in algorithmic error. Further trial and error iteration indicates that the increased algorithmic error in QDNNs stems from initialization and optimization challenges unique to quantum networks. Unlike classical models, QDNNs must respect constraints such as unitarity and entanglement. Our implementation employs strongly entangling layers. While these are designed to capture potential quantum correlations in the data—if present—they also can enhance the model's ability to represent higher-order nonlinearities and interference-like patterns. However, this expressiveness comes at the cost of a more complex and potentially rugged optimization landscape. Another important challenge in training QDNNs is overcoming barren plateaus, where the gradients vanish exponentially as the number of qubits increases. To mitigate this, we used small-angle initialization, sampling parameters from a narrow normal distribution. While this helps to maintain non-vanishing gradients, it also limits the search space and contributes to algorithmic error since it forces the QDNN to explore a constrained region of the parameter space in early training. This can slow down convergence and make the network more sensitive to local minima, leading to greater variability in the extracted CFFs across different replicas, manifesting as higher algorithmic error.  This baseline comparison study suggests that the QDNN extraction can be further improved by refining initialization and optimization techniques and fully leveraging the algorithmic differences between the quantum inspired DNN and the purely classical type. Building upon the insights gained from the Basic QDNN, we will now introduce the Full QDNN, which leverages quantum-specific optimizations to further enhance CFF extraction performance.

\begin{table*}[t!]
    \centering
    \caption{\label{tab:new-QDNNs-compar} The QDNN error contributions are listed for the Basic QDNN, Models 1-3, and Full QDNN for each CFF and DVCS term. The Basic QDNN is the baseline model used for the full extraction in Sec. \ref{sec:CFFExtraction}, Models 1-3 are refined models aimed to take advantage of quantum effects, and the Full QDNN is the optimized QDNN we use for all subsequent extractions.}
    \begin{ruledtabular}
    \begin{tabular}{c|cccc|cccc|cccc|cccc}
        \multirow{2}{*}{Model} & \multicolumn{4}{c|}{Algorithmic Error} & \multicolumn{4}{c|}{Methodological Error} & \multicolumn{4}{c|}{Precision} & \multicolumn{4}{c}{Accuracy} \\
        & $\Re e\mathcal{H}$ & $\Re e\mathcal{E}$ & $\Re e\widetilde{\mathcal{H}}$ & $DVCS$ & $\Re e\mathcal{H}$ & $\Re e\mathcal{E}$ & $\Re e\widetilde{\mathcal{H}}$ & $DVCS$ & $\Re e\mathcal{H}$ & $\Re e\mathcal{E}$ & $\Re e\widetilde{\mathcal{H}}$ & $DVCS$ & $\Re e\mathcal{H}$ & $\Re e\mathcal{E}$ & $\Re e\widetilde{\mathcal{H}}$ & $DVCS$ \\
        \hline
        Basic & 0.174 & 0.241 & 0.241 & 0.00227 & 0.0395 & 0.0482 & 0.0412 & 0.000208 & 0.454 & 0.652 & 1.02 & 0.00520 & 0.759 & 0.0330 & 0.0495 & 0.00259 \\
        1 & 0.135 & 0.169 & 0.170 & 0.00258 & 0.0405 & 0.0468 & 0.0367 & 0.000339 & 0.372 & 0.529 & 0.851 & 0.00638 & 0.765 & 0.0632 & 0.0351 & 0.00252 \\
        2 & 0.147 & 0.191 & 0.196 & 0.000559 & 0.0395 & 0.0471 & 0.0360 & 0.000153 & 0.416 & 0.573 & 0.870 & 0.00498 & 0.752 & 0.0389 & 0.0264 & 0.00255 \\
        3 & 0.137 & 0.177 & 0.173 & 0.00241 & 0.0404 & 0.0469 & 0.0373 & 0.000260 & 0.376 & 0.531 & 0.852 & 0.00533 & 0.764 & 0.0604 & 0.0321 & 0.00260 \\
        Full & 0.127 & 0.161 & 0.175 & 0.000502 & 0.0408 & 0.0470 & 0.0356 & 0.000157 & 0.402 & 0.578 & 0.864 & 0.00511 & 0.773 & 0.0327 & 0.00018 & 0.00264
    \end{tabular}
    \end{ruledtabular}
\end{table*}

\begin{figure}
    \centering
    \includegraphics[width=0.45\textwidth]{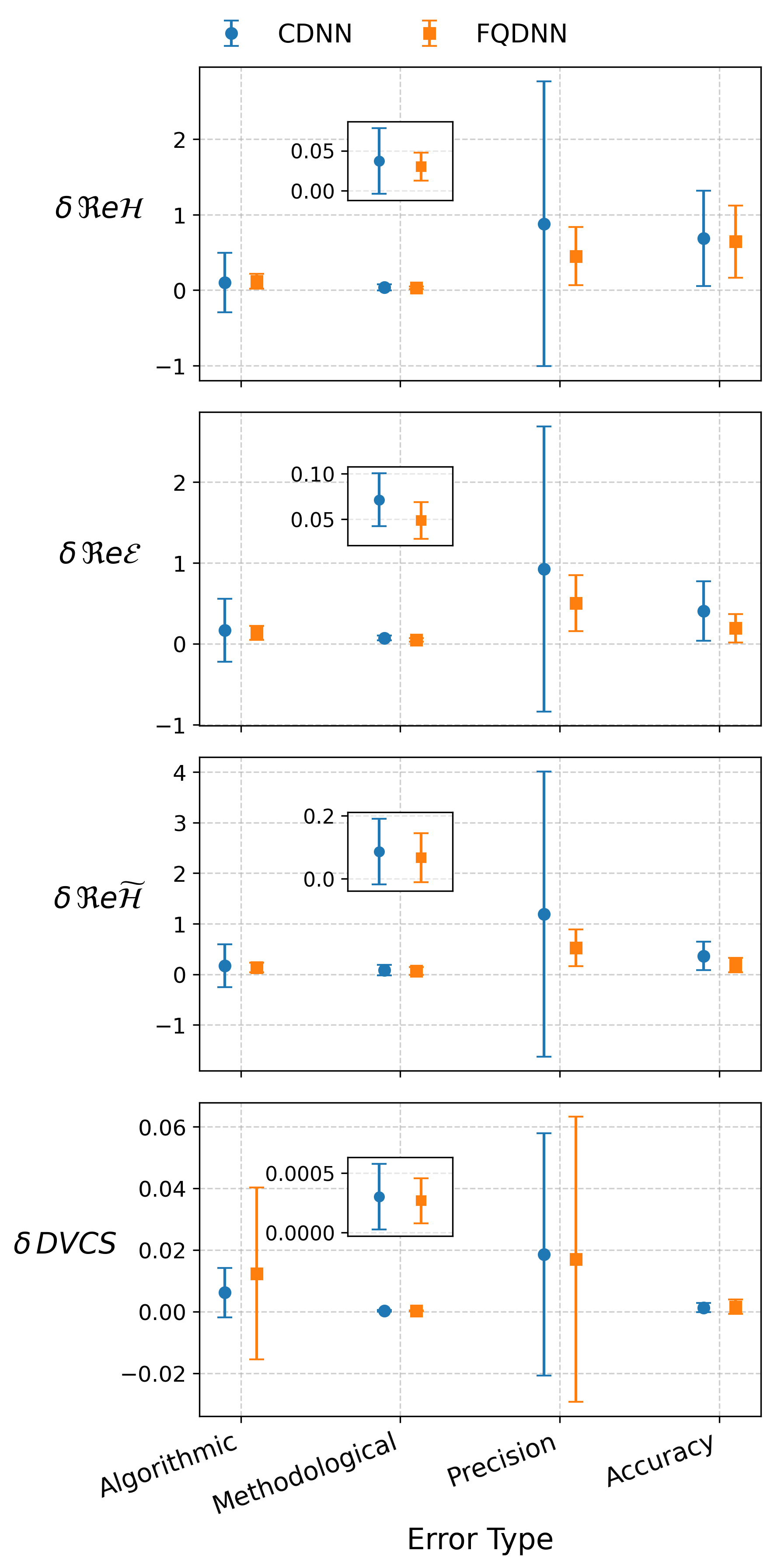}
    \caption{Comparative error distributions of the CDNN (blue) and Full QDNN (orange) extraction across all kinematic bins. The points and error bars show histogram mean and error. These results indicate that QDNNs achieve overall lower errors in extracting CFFs when quantum-specific advantages are leveraged and quantum-specific challenges are mitigated.}
    \label{fig:FQDNN_CFF_error_results}
\end{figure}

The observed differences in performance between the QDNN and CDNN architectures are driven by a combination of inductive bias, functional complexity, and data characteristics. The unitary structure and comparatively reduced parameterization of the quantum circuit layer constrain the hypothesis space, which can act as an implicit regularizer in bins where the $\phi$-dependent structure is weakly constrained by the data. In particular, when the azimuthal dependence is close to linear or dominated by a small number of harmonic components, and when experimental uncertainties are large or the sampling in $\phi$ is sparse, a lower-capacity model may suppress overfitting to statistical fluctuations. Conversely, in bins where the cross-section exhibits more complex interference structure or stronger nonlinear dependence on the underlying CFF combinations, higher expressive capacity may be beneficial.

In addition, the QDNN employs a distinct feature mapping based on angle embedding and entangling layers, which generates nonlinear correlations in a manner different from standard dense layers. This altered representation can modify the effective optimization landscape. However, trainability considerations—such as circuit depth, initialization strategy, and entanglement range—also play a significant role. In the next section the ``Full QDNN'' modifications introduced are designed to mitigate barren-plateau behavior and improve gradient flow, thereby stabilizing convergence.

Because both signal complexity and experimental signal-to-noise ratios vary bin-by-bin, no uniform performance hierarchy is expected across all kinematics. The quantum qualifier $\widehat{\Xi}_{\mathrm{DVCS}}$ is therefore introduced as a pragmatic, per-bin model-selection heuristic rather than an assertion of universal quantum advantage.

\subsection{Fully Optimized QDNN}
\label{subsec:leveraging-quantum}

In this section, we develop a new QDNN, which we refer to as the Full QDNN (FQDNN), that aims to optimize CFF extraction by employing quantum-specific capabilities and mitigating quantum-specific challenges (Sec. \ref{subsubsec:fqdnn}). Then, we use the FQDNN to perform the same error analysis as in Sec. \ref{subsec:bqdnn} and briefly discuss the results (Sec. \ref{subsubsec:fqdnn-extraction}).

\subsubsection{Full QDNN Model}
\label{subsubsec:fqdnn}

To systematically explore quantum-specific optimizations, we explore three new QDNN models (Models 1-3), each incorporating targeted modifications designed to enhance specific aspects of the architecture with the goal of improving stability and reducing errors. Our baseline model, the Basic QDNN—introduced earlier in Sec.~\ref{subsec:bqdnn}—is a straightforward quantum adaptation of the CDNN. Although not fully optimized, it serves as a proof of concept for achieving a quantum advantage in CFF extraction and provides a valuable foundation for identifying potential improvements. The QDNN algorithmic error improvements observed from Models 1-3 are more comparable to the algorithmic errors from the CDNN extraction (see Fig.~\ref{fig:CFF_error_results}) and suggest key strategies for reducing algorithmic error and improving precision in quantum deep learning models for CFF extraction. Specifically, by refining the entanglement structure, implementing layer-wise training, and adjusting initialization schemes, we demonstrate that targeted modifications can significantly enhance training stability and overall model reliability. Therefore, to construct our FQDNN, we incorporate all the improvements from Models 1-3, including tunable entanglement range (Model 1), layer-wise training (Model 2), and depth-scaled, larger-angle initialization (Models 2 and 3), while maintaining the baseline architecture from the Basic QDNN, including a width of 6 qubits and classical pre- and post-processing layers.

For clarity, we provide a detailed description of the architecture of the FQDNN as follows. The classical preprocessing layer is a single fully-connected linear layer that maps the 3-dimensional input \((x_B, Q^2, t)\) to a 6-dimensional vector, matching the number of qubits in the quantum device \citep{paszke2019pytorch}. The quantum circuit then encodes this input using angle embedding, which applies rotation gates parameterized by input features \citep{schuld2015introduction}. Following encoding, the circuit applies strongly entangling layers, where entanglement is controlled by a new entanglement strength hyperparameter \citep{Farhi2018}. We choose nearest-neighbor entanglement for the FQDNN \citep{Cerezo2021}. 

Trainable weights in the quantum circuit are initialized with a depth-scaled scheme: the standard deviation of the normal distribution used for initialization is scaled as \(1/\sqrt{l+1}\) for the \(l\)-th layer \citep{glorot2010understanding}. This encourages smaller initial angles in deeper layers, which mitigates barren plateaus by ensuring gradients are initially non-zero and preventing over-saturation \citep{mcclean2018barren}. The output of the quantum circuit consists of the expectation values \(\langle Z \rangle\) for each qubit, forming a 6-dimensional vector \citep{schuld2015introduction}. 

This vector is then passed through a classical postprocessing layer consisting of a fully-connected linear layer with 64 hidden units and Tanh activation, followed by a final output layer mapping to the four CFFs \citep{paszke2019pytorch}. The quantum circuit is dynamically rebuilt as the depth of the circuit gradually increases from 2 layers to 8 layers, using a factory function that generates a new quantum node with the updated number of layers \citep{Cerezo2021}.

To test the FQDNN, we perform the same error analysis used on Models 1-3 and display the results in Table \ref{tab:new-QDNNs-compar}. Incorporating features from Models 1-3 allows the FQDNN to perform better overall than each of the individual models, with improved algorithmic error and precision over all 4 CFFs. With these example kinematics, the FQDNN also exhibits improved methodological error for $\Re e\mathcal{E}$, $\Re e\widetilde{\mathcal{H}}$, and $DVCS$ and improved accuracy for $\Re e\widetilde{\mathcal{H}}$ and $DVCS$. 

Model 1 modifies the strongly entangling layers by adding an entanglement range parameter, limiting interactions to nearest neighbors in some cases rather than fully entangling all qubits at every layer. This aims to balance expressivity and stability, reducing excessive parameter correlations that can destabilize optimization. Model 2 focuses on addressing the barren plateau problem, a QDNN-unique issue characterized by gradients that vanish exponentially as the circuit depth increases. To mitigate this, Model 2 employs layerwise training, where the network begins with a shallow quantum circuit and progressively increases the number of layers over training. Additionally, it introduces depth-scaled initialization, ensuring that each layer’s parameter magnitudes decrease as depth increases. This prevents excessive parameter magnitudes in deeper layers while maintaining trainable gradients throughout optimization. Model 3 refines the small-angle initialization used in the Basic QDNN, which is already known to help prevent barren plateaus at initialization but may overly restrict the search space, leading to increased error. Instead of reducing the initial weight magnitudes across layers, Model 3 rescales small-angle initialization to maintain better trainability without completely restricting parameter exploration.

To test these new models, we perform the CFF extraction on 1000 noisy replicas of Set 144 (see Fig.~\ref{fig:ReE-dist}) using each of these four models and display the algorithmic error results in Table \ref{tab:new-QDNNs-compar}. Comparing the results of the Basic QDNN with Fig.~\ref{fig:CFF_error_results}, we observe that the errors for this kinematic bin align relatively well with the average errors of the extracted CFFs across all bins. Performing the extraction with the improved QDNN models, the algorithmic error and precision improve significantly for all three models. Table \ref{tab:new-QDNNs-compar} indicates that Model 1 produces the lowest algorithmic errors for $\Re e\mathcal{H}$, $\Re e\mathcal{E}$, and $\Re e\widetilde{\mathcal{H}}$, while Model 2 produces the lowest algorithmic error for $DVCS$. The improved precision follows a similar pattern. In this example, we find that these new models significantly improve accuracy and methodological error for $\Re e\widetilde{\mathcal{H}}$, but there is no overall trend for these two metrics.

\begin{figure}[ht!]
    \centering
    \subfloat[CDNN]{\includegraphics[width=0.45\textwidth]{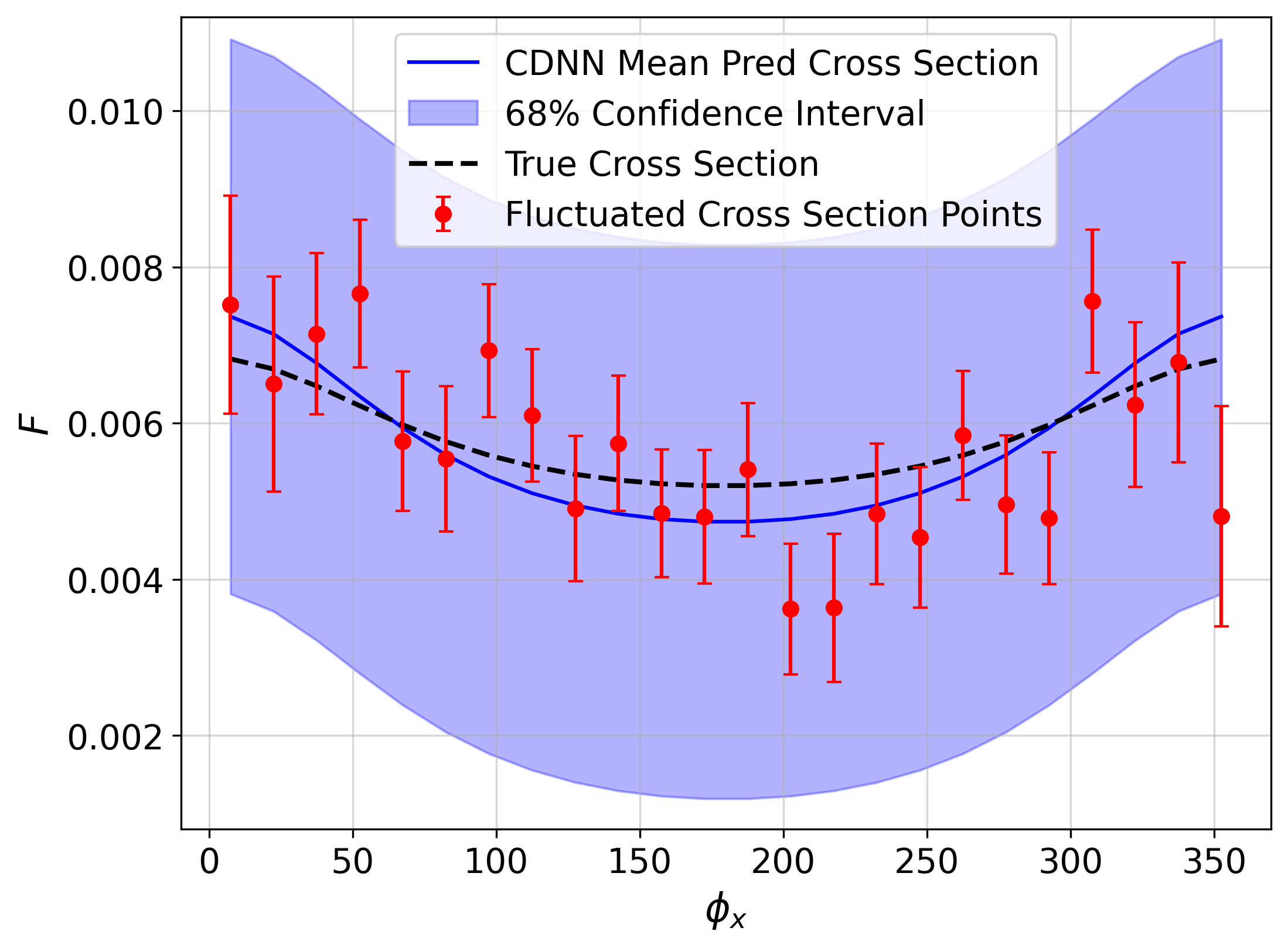}\label{fig:c-fvsphi-all}} \\
    \subfloat[Basic QDNN]{\includegraphics[width=0.45\textwidth]{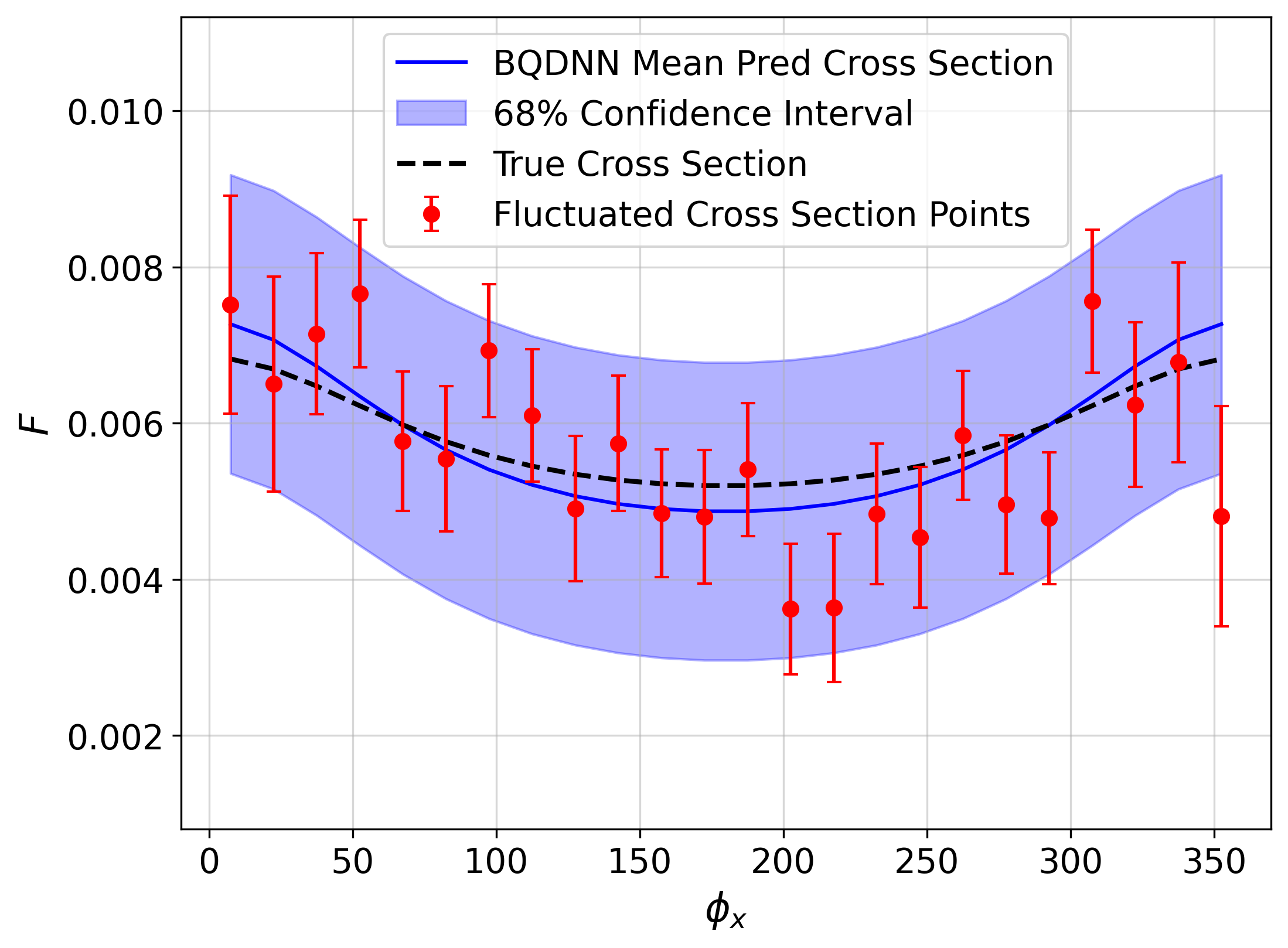}\label{fig:bq-fvsphi-all}} \\
    \subfloat[Full QDNN]{\includegraphics[width=0.45\textwidth]{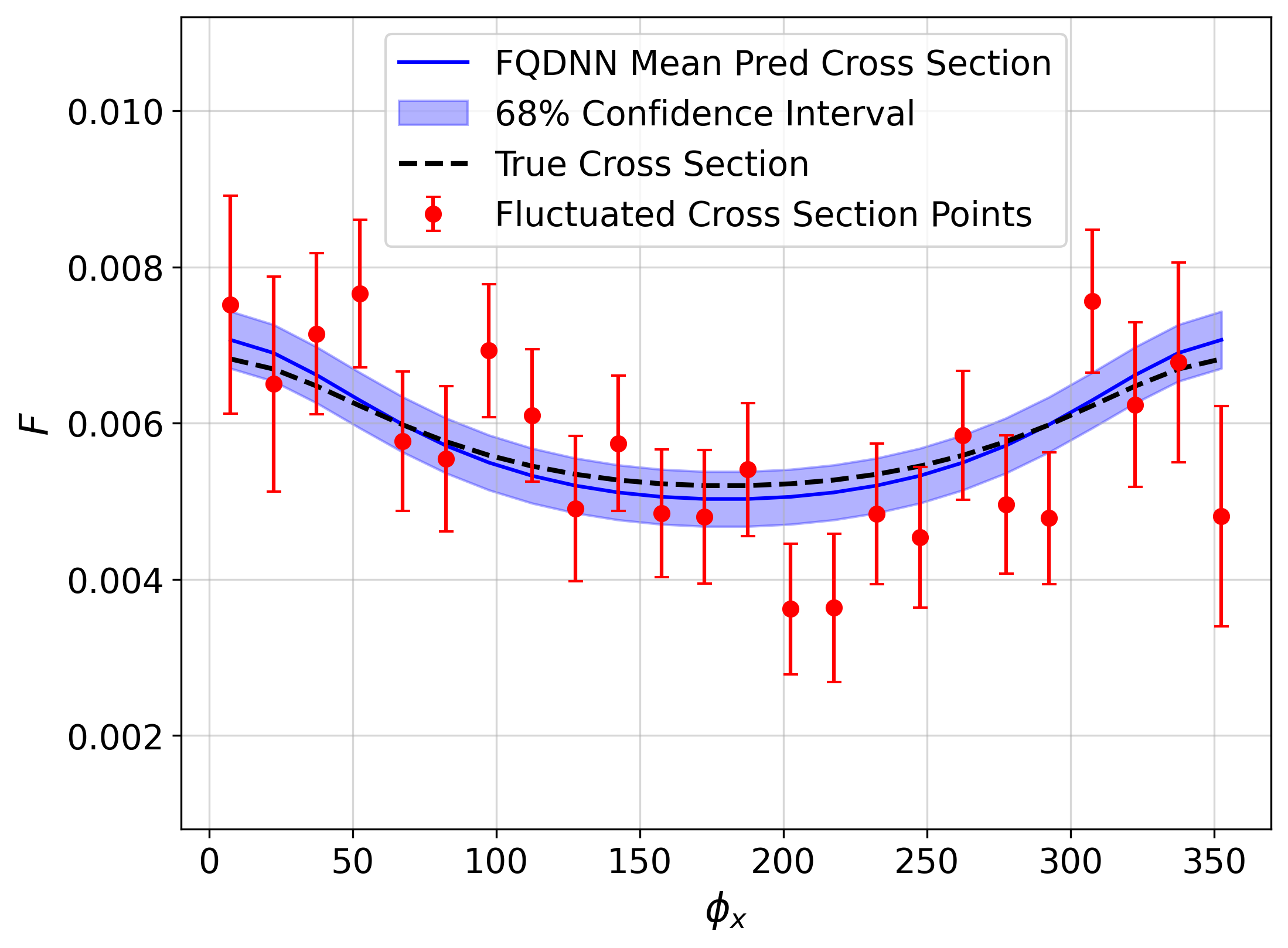}\label{fig:fq-fvsphi-all}} \\
    \caption{Predicted cross sections from the (a) CDNN, (b) Basic QDNN, and (c) Full QDNN fits of noisy replicas of cross section pseudodata (red points) generated from kinematics $k=8.847$ GeV, $Q^2 = 5.36$ GeV$^2$, $x_B = 0.485$, $t = 0.50851$ GeV$^2$ (Set 48). In this example, the FQDNN exhibits a dramatic improvement from the CDNN and BQDNN in terms of proximity to the true cross section and a small improvement in accuracy.}
    \label{fig:fvsphi-all}
\end{figure}

\subsubsection{CFF Extraction Results}
\label{subsubsec:fqdnn-extraction}

We now summarize our results of the full CFF extraction from the pseudodata sets using the CDNN and FQDNN. The results of our error propagation analysis are summarized in Fig.~\ref{fig:FQDNN_CFF_error_results}. These results demonstrate overall improved FQDNN performance in terms of methodological error and accuracy, and superior FQDNN performance in terms of precision. The CDNN and FQDNN perform similarly in terms of algorithmic error. It is also clear that the DVCS term does not benefit from the FQDNN in comparison to the Basic QDNN. However, in terms of the predicted cross section, it is clear from the example in Fig.~\ref{fig:fvsphi-all} that the FQDNN performs better than both the CDNN and BQDNN overall.


These results demonstrate that QDNNs provide a powerful framework for extracting
CFFs from DVCS data and, when the quantum architecture is fully optimized,
generally achieve smaller uncertainties than CDNNs. The results shown in
Figs.~\ref{fig:CFF_error_results} and~\ref{fig:FQDNN_CFF_error_results} represent
means evaluated over the full kinematic range of the data. While the FQDNN
typically yields CFFs that better reproduce the true cross section, the CDNN can
outperform the QDNN in certain kinematic regions or error regimes. This naturally
raises the question of how to select the optimal algorithm for a given dataset,
kinematic configuration, or experimental uncertainty. To address this, we
introduce a novel \emph{quantum qualifier} metric that predicts \emph{a priori}
whether a CDNN or FQDNN is better suited for CFF extraction under specified
experimental conditions by identifying regimes in which QDNNs are expected to
outperform CDNNs.

\begin{figure*}[htp!]
    \centering
    \subfloat[$0\sigma_F$ experimental error]{\includegraphics[width=0.45\textwidth]{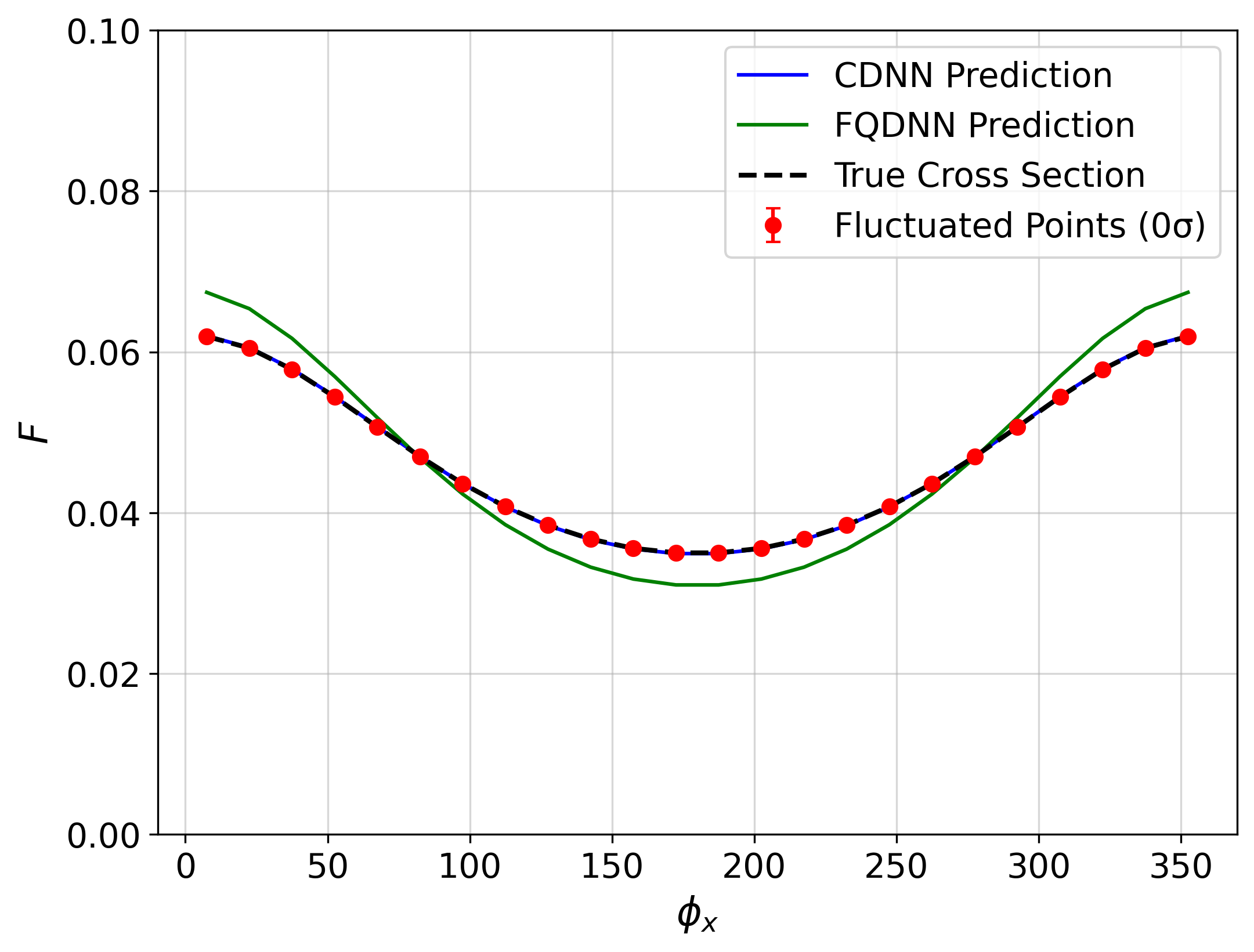}\label{fig:comparison_fit_0sigma}} 
    \hfill
    \subfloat[$2\sigma_F$ experimental error]{\includegraphics[width=0.45\textwidth]{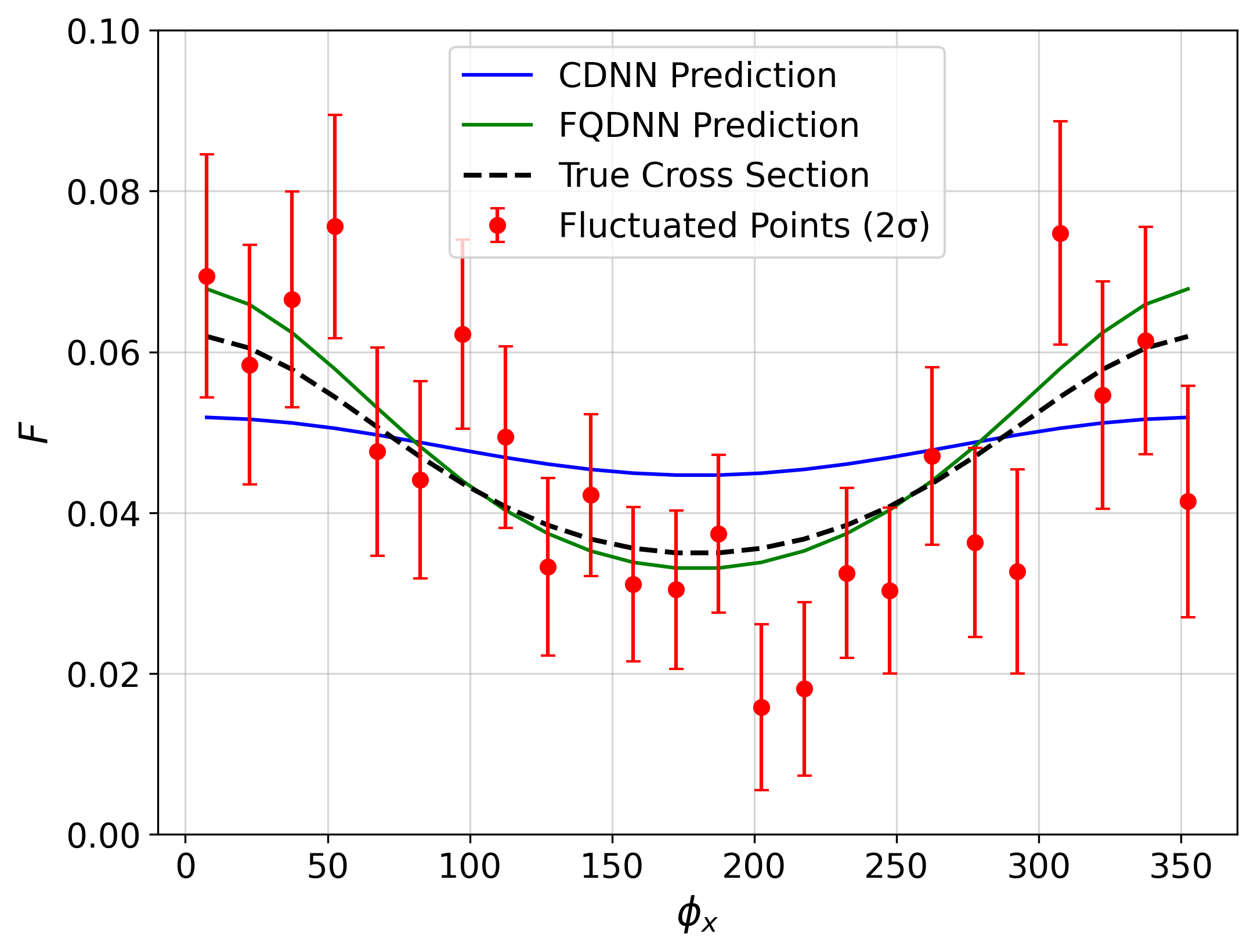}\label{fig:comparison_fit_2sigma}} 
    \caption{Predicted cross section from the CDNN and QDNN fits of cross section pseudodata (red points) generated from kinematics $k=5.75$ GeV, $Q^2 = 2.012$ GeV$^2$, $x_B = 0.378$, $t = -0.192$ GeV$^2$ and CFFs $\Re e\mathcal{H}\approx -0.808$, $\Re e\mathcal{E}\approx -1.229$, $\Re e\widetilde{\mathcal{H}}\approx -1.360$, and $DVCS = 0.0114$. This example comparison illustrates the general trend of increased quantum outperformance $\Xi_{DVCS}$ with greater experimental error.}
    \label{fig:QQ-comparison-fit}
\end{figure*}

\section{DVCS Quantum Qualifier}
\label{sec:dvcs-qq}

In this section, we construct a quantum qualifier $\hat{\Xi}_{DVCS}$ tailored specifically for DVCS data. The goal of the quantum qualifier is to determine whether the CDNN or FQDNN (described in Sec. \ref{subsec:leveraging-quantum}) should be used for the extraction of CFFs from a given set of experimental DVCS data, based solely on its functional characteristics and error. The quantum qualifier is designed based on CDNN and FQDNN extractions from systematically constructed pseudodata with physically realistic kinematics, CFFs, and error. Then, the quantum qualifier is tested, and in the following section (Sec. \ref{sec:cff-extract-exp-data}), it is used to perform a preliminary check to determine which algorithm (QDNN or CDNN) is optimal for that particular kinematic set for maximal extraction of CFF information. 

\subsection{Pseudodata Generation}
\label{subsec:qq-pseudo-gen}

To construct the DVCS quantum qualifier, we generate a controlled ensemble of pseudodata based on the kinematic values of the experimental measurements. As a first step, we perform a baseline CDNN extraction on 1000 replicas of each experimental data set, sampling each cross-section point within its quoted uncertainty. This provides a physically grounded range of CFF values—typically between $-5$ and $5$—that serves as the foundation for the pseudodata construction.

Using these CDNN-extracted CFFs, we systematically generate pseudodata for each of the 195 kinematic sets. For every CFF, we select five values evenly spaced within the 95\% confidence interval obtained from the baseline extraction, and for each cross-section point we impose one of four error scalings: $0$, $0.5\sigma_F$, $1\sigma_F$, or $2\sigma_F$, where $\sigma_F$ is the experimental uncertainty. This results in $5^4 \cdot 4 = 2500$ pseudodata replicas per kinematic point, providing a comprehensive and physically consistent dataset for training the DVCS quantum qualifier.

\subsection{Quantum Qualifier}
The DVCS quantum qualifier $\hat{\Xi}_{DVCS}$ is designed to estimate the underlying quantum outperformance metric $\Xi_{DVCS}$. Defining this metric requires the quantity $M_{DVCS}$, given by:
\begin{equation}
    M_{DVCS} = \int_{\phi_{\text{min}}}^{\phi_{\text{max}}}|F_{\text{DNN}}(\phi) - F_{\text{true}}(\phi)|d\phi,
\end{equation}
where $F_{\text{DNN}}(\phi)$ is the predicted cross section curve from the DNN-extracted CFFs and $F_{\text{true}}(\phi)$ is the true cross section curve (with no experimental error). Most of the pseudodata sets have $\phi_{\text{min}} = 7.5^\circ$ and $\phi_{\text{max}} = 352.5^\circ$. Then, the quantum outperformance $\Xi_{DVCS}$ is defined as 
\begin{equation}
    \Xi_{DVCS} = M_{DVCS}^{\text{CDNN}} / M_{DVCS}^{\text{QDNN}} - 1,
\end{equation}
so $\Xi_{DVCS}>0$ means that the QDNN achieves a better fit, and vice versa. The magnitude of $\Xi_{DVCS}$ indicates the extent to which the better DNN outperforms.

We perform the CFF extraction on the pseudodata sets using the CDNN and FQDNN, which gives us a quantum out-performance value for each of the 2500 replicas of each of the 195 sets of kinematics. We then examine the $\Xi_{DVCS}$-dependence on five function characteristics: nonlinearity, frequency complexity, fractal dimension, mutual information, and Fourier transform complexity. We construct a simple expression that serves as a useful indicator for which DNN to use, so we only consider the strongest correlations. Additionally, since both the CDNN and FQDNN are optimized, the quantum qualifier $\hat{\Xi}_{DVCS}$ is not dependent on the number of epochs $n$. We find that $\Xi_{DVCS}$ has the strongest correlation with nonlinearity ($\mathfrak{N}$) \cite{montgomery2012}, and obtain the relation from out fit analysis $\Xi_{DVCS}=-1.19\mathfrak{N}+0.65$. This indicates that a higher nonlinearity results in lower QDNN out-performance.

Nonlinearity ($\mathfrak{N}$) measures the deviation of a dataset $(x_i,y_i)$ from a linear trend and can reveal the expressive power of a model in capturing nonlinear relationships. To compute it, we first perform a linear regression on the set of data points $(x_i,y_i)$, which yields points $(x_i, \hat{y}_i^{\text{lin}})$. Then, the nonlinearity metric is defined to be the ratio between the residual sum of squares of the linear fit and the residual sum of squares from the mean:
\begin{equation}
    \mathfrak{N} = \frac{\sum_{i=1}^{n} (y_i - \hat{y}_i^{\text{lin}})^2}{\sum_{i=1}^{n} (y_i - \bar{y})^2},
\end{equation}
where 
\begin{equation}
    \bar{y} = \frac{1}{n}\sum_{i=1}^{n} y_i.
\end{equation}
Values of $\mathfrak{N}$ near 0 indicate the data is well-approximated by a linear model, while values closer to 1 suggest higher nonlinearity. In the quantum setting, parameterized quantum circuits have been shown to exhibit enhanced nonlinear representational capacity. Du \textit{et al.} \cite{du2020expressive} demonstrate that QDNNs possess a richer functional landscape than classical models due to quantum operations. However, as defined here, the performance metric has a negative correlation with nonlinearity.

The magnitude of the experimental error significantly influences the potential for quantum advantage. To quantify this in the qualifier, we examine the explicit dependence of $\Xi_{DVCS}$ on the relative error. We define the average scaled error $\overline{\epsilon}_s$ as the mean of the ratio $s\sigma_{F,i}/F_{i,r}$ over all valid kinematic points $i$ and replicas $r$, where $s$ is the error scaling factor (e.g., $s = 0, 0.5, 1, 2$), $\sigma_{F,i}$ is the experimental uncertainty at point $i$ (identical across replicas), and $F_{i,r}$ is the corresponding cross section value:
\begin{equation}
\overline{\epsilon}_s(k,Q^2,x_B,t) = \frac{1}{N}\sum_{i,r}\frac{s\sigma_{F,i}}{F_{i,r}},
\end{equation}
with $N$ the total number of $(i,r)$ pairs. Varying $\overline{\epsilon}_s$ across the pseudodata sets from 0 to 0.6 allows us to study the impact of increasing experimental uncertainty. We find a strong positive correlation between $\Xi_{DVCS}$ and $\overline{\epsilon}_s$, well-approximated by the linear relation $\Xi_{DVCS} = 2.00\overline{\epsilon}_s - 0.66$.

This trend is illustrated in Fig.~\ref{fig:QQ-comparison-fit}. When there is no experimental error, the CDNN achieves a nearly perfect fit (Fig.~\ref{fig:comparison_fit_0sigma}), whereas the FQDNN underperforms slightly. However, under high experimental uncertainty (Fig.~\ref{fig:comparison_fit_2sigma}), the CDNN performance degrades significantly, while the FQDNN remains robust—highlighting the increased quantum outperformance with increasing error.

The DVCS quantum qualifier, combining $\Xi_{DVCS}$ dependencies on $\overline{\epsilon}_s$ and $\mathfrak{N}$ weighted by correlation strength, is given by:
\begin{equation}
    \hat{\Xi}_{DVCS} = 1.98\overline{\epsilon}_s - 0.132\mathfrak{N} - 0.583.
    \label{eq:dvcs-qq}
\end{equation}
A positive $\hat{\Xi}_{DVCS}$ suggests the QDNN will likely outperform, with its magnitude indicating the strength of correlation with out-performance. While $\hat{\Xi}_{DVCS}$ guides predictions of QDNN advantage, it does not precisely determine $\Xi_{DVCS}$, which depends on data structure and kinematics. A linear fit of $\Xi_{DVCS} = 1.0004\hat{\Xi}_{DVCS} - 0.0042$ ($R^2=0.42$), based on average $\hat{\Xi}_{DVCS}$ across kinematic sets and noise level $s$ (excluding extreme outliers), confirms approximate agreement with quantum out-performance.

The relative performance of the QDNN and CDNN exhibits a clear dependence on 
the underlying structure of the DVCS cross section in each local kinematic 
region.  As the functional form of the cross section varies across 
$(x_B,Q^2,t)$ for a set of points in $\phi$, so does its degree of curvature and oscillatory behavior.  
The nonlinearity measure $\mathfrak{N}$ provides a quantitative proxy for this 
local functional complexity: regions with small $\mathfrak{N}$ are effectively 
linear, while regions with large $\mathfrak{N}$ exhibit pronounced curvature or 
higher–harmonic content.

In addition to function nonlinearity, the statistical structure of the data plays an 
equally important role.  The scale of the error in the cross section 
$\overline{\epsilon}_s$ characterizes the local signal-to-noise ratio, and is strongly 
correlated with the reliability of the learned mapping as is the data sparsity.  In regions where the 
data are sparse or dominated by large statistical fluctuations, CDNNs tend to 
overfit individual noisy points because they possess many unconstrained degrees 
of freedom.  The QDNN, by contrast, benefits from the implicit regularization 
introduced by unitary transformations, which suppresses overfitting and yields 
smoother interpolations when information is limited.  As a result, sparse or 
high-noise data can favor the QDNN even when the cross section is only moderately 
nonlinear.  Taken together, the combined dependence on nonlinearity 
$\mathfrak{N}$, statistical variation $\varepsilon_s$, and the kinematics of the 
data uniquely determines which architecture is favored. 
Sparsely sampled kinematic regions with significant statistical noise also tend to 
yield a quantum advantage.  In terms of kinematics regarding the dataset in hand, the CDNN is most likely to perform better in the lower $x$, low $|t|$ and lower $Q^2$ regions even when sparsity and errors are equal.

This simple utility provides the means of checking experimental data sets to determine which algorithm type would perform best and can be used to determine the optimal fit strategy for any given set of experimental data. Note that $\hat{\Xi}_{DVCS}$ is a selection heuristic, not a performance guaranty.  For our pseudodata set the conclusion made by using this heuristic fails on less than 3\% of the test data. With the quantum qualifier defined and its predictive power established, we apply our optimized CFF extraction method to real experimental data for improved extraction.

\section{Optimized CFF Extraction from Experimental Data}
\label{sec:cff-extract-exp-data}

This section demonstrates the practical application of the DVCS quantum qualifier $\hat{\Xi}_{DVCS}$ for optimized CFF extraction on a subset of real experimental data (see Sec. \ref{sec:ExperimentalData}), validating our QDNN approach in a realistic setting beyond pseudodata. We select a subset where $\hat{\Xi}_{DVCS}$ predicts quantum outperformance in most kinematics to evaluate the quantum advantage.  This subset has enough data for a good demonstration, but with a variety of experimental errors.

In Sec. \ref{subsec:realistic-pseudo-gen}, we describe generating realistic pseudodata from the initial CDNN extraction of experimental data (Sec. \ref{subsec:qq-pseudo-gen}). In Sec. \ref{subsec:realistic-pseudo-extract}, we tune CDNN and FQDNN on this pseudodata to extract physically realistic CFFs and assess the DVCS quantum qualifier’s effectiveness. Finally, in Sec. \ref{subsec:exp-data-extract}, we perform a $\hat{\Xi}_{DVCS}$-guided CFF extraction on the selected experimental data subset and present the results.

\begin{table}[tb!]
\caption{\label{tab:table_par_cff_realistic}%
The parameters used in Eq.~\eqref{equ:model-par} to generate the realistic pseudodata set, fitted from the initial CDNN extraction in Sec. \ref{subsec:qq-pseudo-gen}.
}
\begin{ruledtabular}
\begin{tabular}{ccccccc}
CFFs & $a$ & $b$ & $c$ & $d$ & $e$ & $f$ \\
\hline
$\Re e \mathcal{H}$ & -8.13 & 1.82 & 35.26 & 25.37 & 6.26 & 3.20 \\
$\Re e \mathcal{E}$ &  6.92 & -5.64 & 0.81 & 0.98 & 4.03 & 49.71   \\
$\Re e \widetilde{\mathcal{H}}$ & -8.51 & 1.72 & 31.11 & 22.49 & 6.09 & 4.77  \\
$DVCS$ & 0.45 & -0.45 & 4.40 & 2.91 & 0.13 & 0.08 \\
\end{tabular}
\end{ruledtabular}
\end{table}

\begin{figure}[htp!]
    \centering
    \subfloat[$\Re e\mathcal{H}$ vs. $-t$ with $x_B=0.275$]{\includegraphics[width=0.45\textwidth]{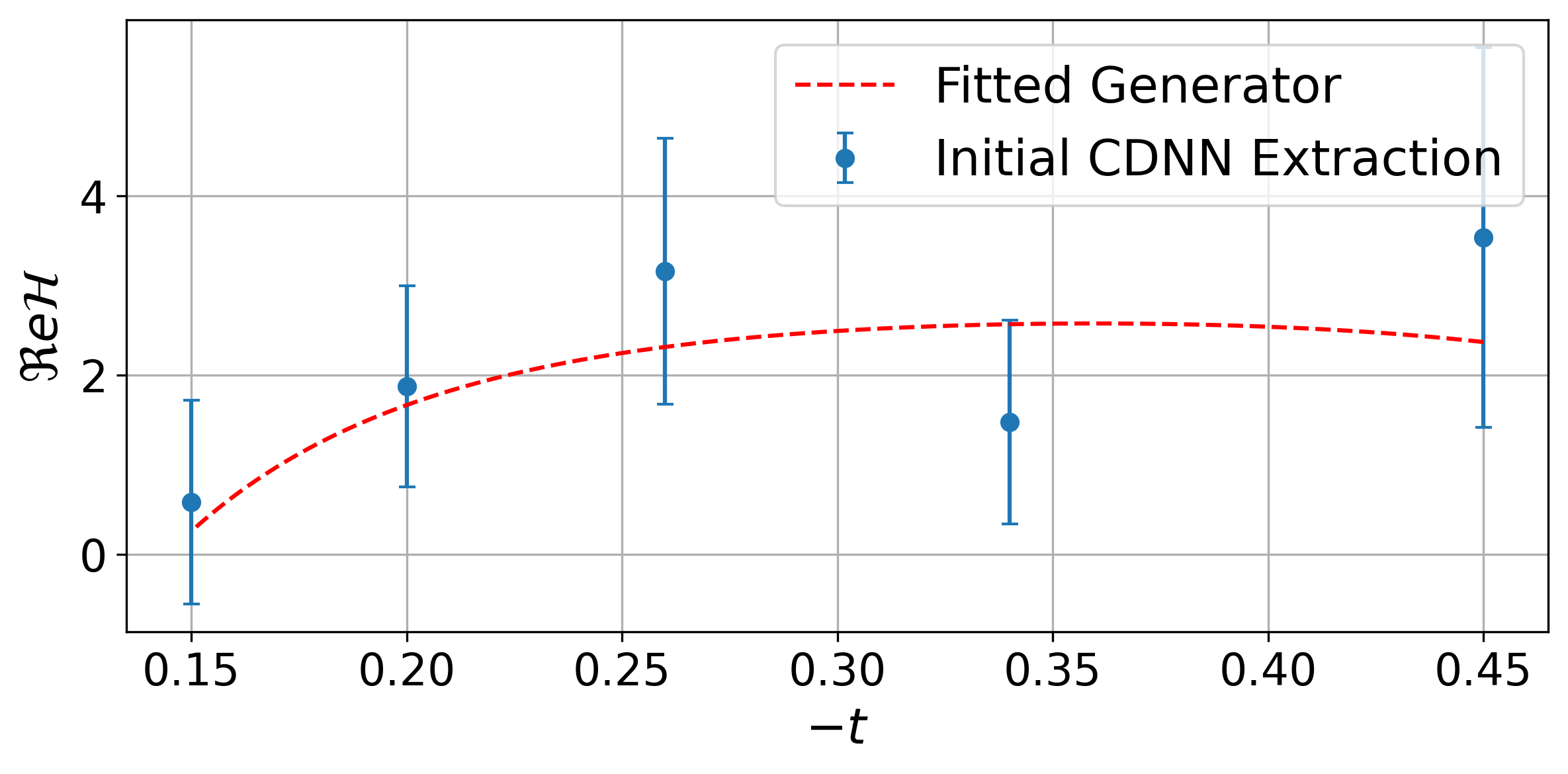}\label{fig:fit-gen-ReH}} \\[0.5cm]
    \subfloat[$\Re e\mathcal{E}$ vs. $-t$ with $x_B=0.365$]{\includegraphics[width=0.45\textwidth]{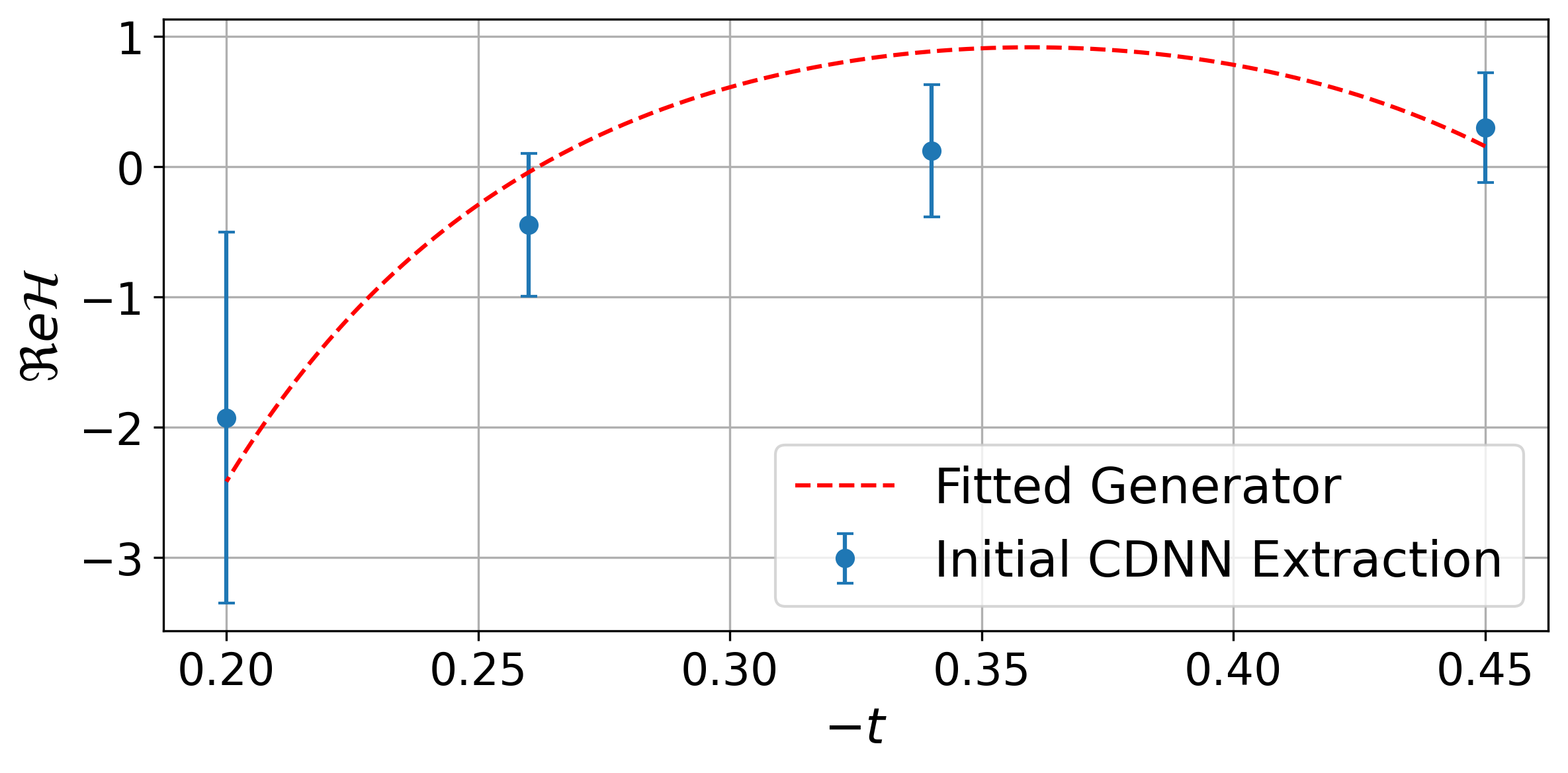}\label{fig:fit-gen-ReE}} \\[0.5cm]
    \subfloat[$\Re e\widetilde{\mathcal{H}}$ vs. $-t$ with $x_B=0.399$]{\includegraphics[width=0.45\textwidth]{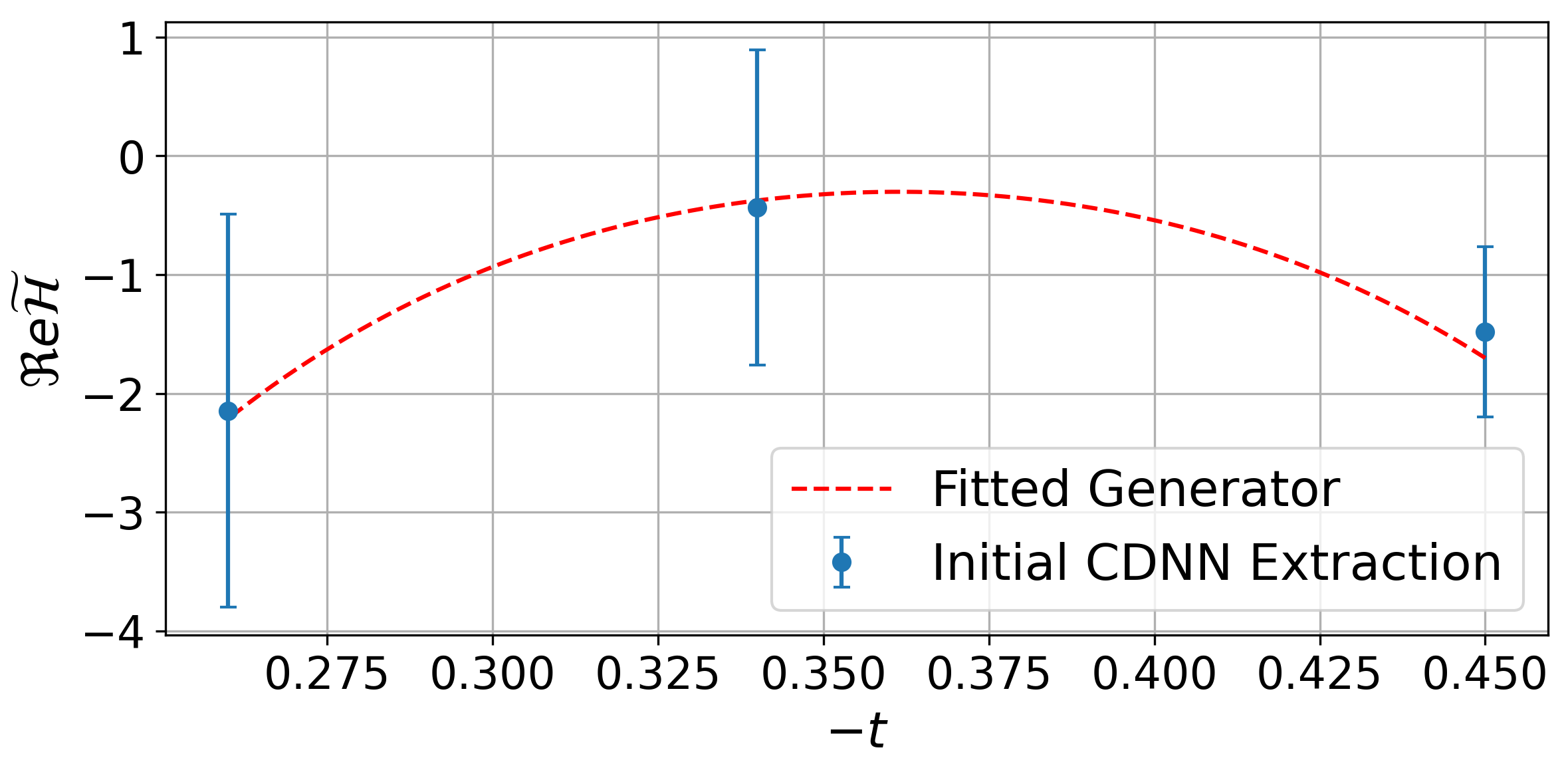}\label{fig:fit-gen-ReHt}} \\[0.5cm]
    \subfloat[$DVCS$ vs. $-t$ with $x_B=0.244$]{\includegraphics[width=0.45\textwidth]{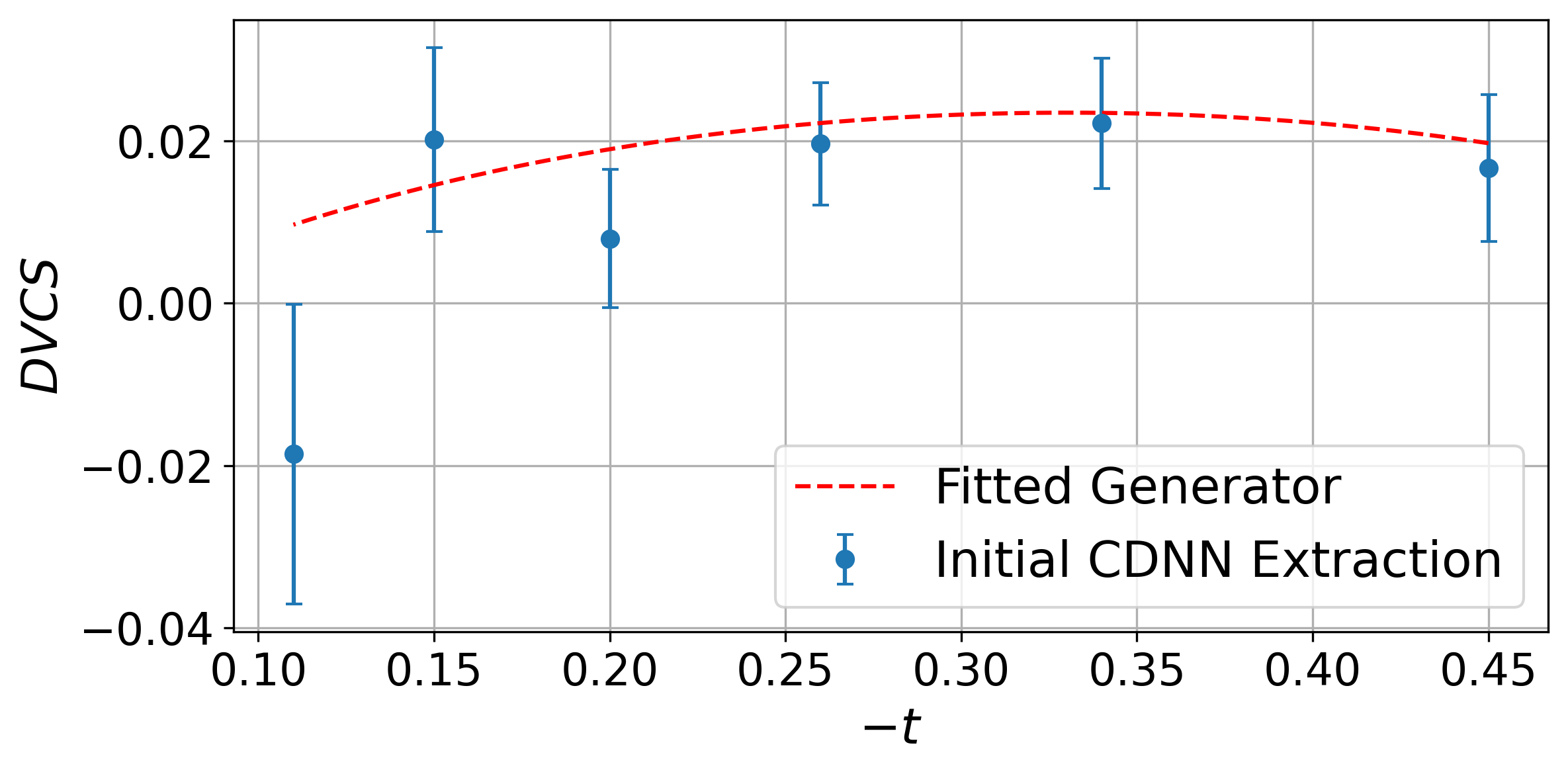}\label{fig:fit-gen-dvcs}}
    \caption{Example comparisons between the initial CDNN-extracted CFFs (blue with error bars) and the CFFs produced from the generating function [Eq.~\eqref{equ:model-par}] using the fitted parameters in Table \ref{tab:table_par_cff_realistic} (red). The CFFs are plotted against $-t$ for select values of $x_B$ in the subset of all experimental data specified in the text.}
    \label{fig:fitted-generator-examples}
\end{figure}

\subsection{Realistic Pseudodata Generation}
\label{subsec:realistic-pseudo-gen}

We choose to examine a specific subset of the experimental data (Sets 108 to 175) that has kinematics $k=5.75$ GeV, $Q^2$ from 1.79 GeV$^2$ to 3.77 GeV$^2$, $x_B$ from $0.244$ to $0.475$, and $t$ from $-0.11$ GeV$^2$ to $-0.45$ GeV$^2$. This subset selects 1145 points in a physically relevant region of phase space covered by JLab experiments (Sec. \ref{sec:ExperimentalData}), where the DVCS cross-section data has a range of experimental errors.

Realistic pseudodata is required for the testing and optimization phase.
In order to generate the pseudodata in this range of kinematics, we begin by modeling the underlying CFFs as smooth functions of the kinematic variables $x_B$ and $t$. Specifically, we fit the parameters $a,b,c,d,e,f$ of the \textit{basic} generating function, Eq.~\eqref{equ:model-par}, to the CFFs obtained from the initial CDNN extraction described in Sec. \ref{subsec:qq-pseudo-gen}, which provides pointwise estimates of the true CFFs at each kinematic setting. The fit is performed using a non-linear least squares method that minimizes the squared difference between the predicted values of the generating function and the CDNN-extracted CFFs across all kinematic points in the subset. The resulting optimized parameter values are shown in Table \ref{tab:table_par_cff_realistic}. To validate the quality of the fit, Fig.~\ref{fig:fitted-generator-examples} compares the fitted generating function predictions with the CDNN-extracted CFFs, which demonstrates that the fitted functions capture the general trend and magnitude of each CFF across the kinematic region.

Using the improved generating function as a stand-in for the ``true'' kinematic dependence of the CFFs, we construct a new set of realistic pseudodata. Following the procedure in Sec. \ref{sec:ExperimentalData}, we use the fitted function to generate the CFFs for each experimental kinematic setting. These CFFs are then used to compute the DVCS cross section at each $\phi$ point. To simulate realistic measurement uncertainties, we add Gaussian noise to each cross-section value based on the established experimental error. Finally, for each kinematic point, we generate 1000 replicas of noisy cross-section data by resampling within the experimental uncertainty. These realistic pseudodata sets are then used for the tuning and evaluation of the DNNs.

\begin{figure}[htbp!]
    \centering
    \subfloat[$\Re e\mathcal{H}$ vs $-t$ with $x_B=0.244$]{\includegraphics[width=0.375\textwidth]{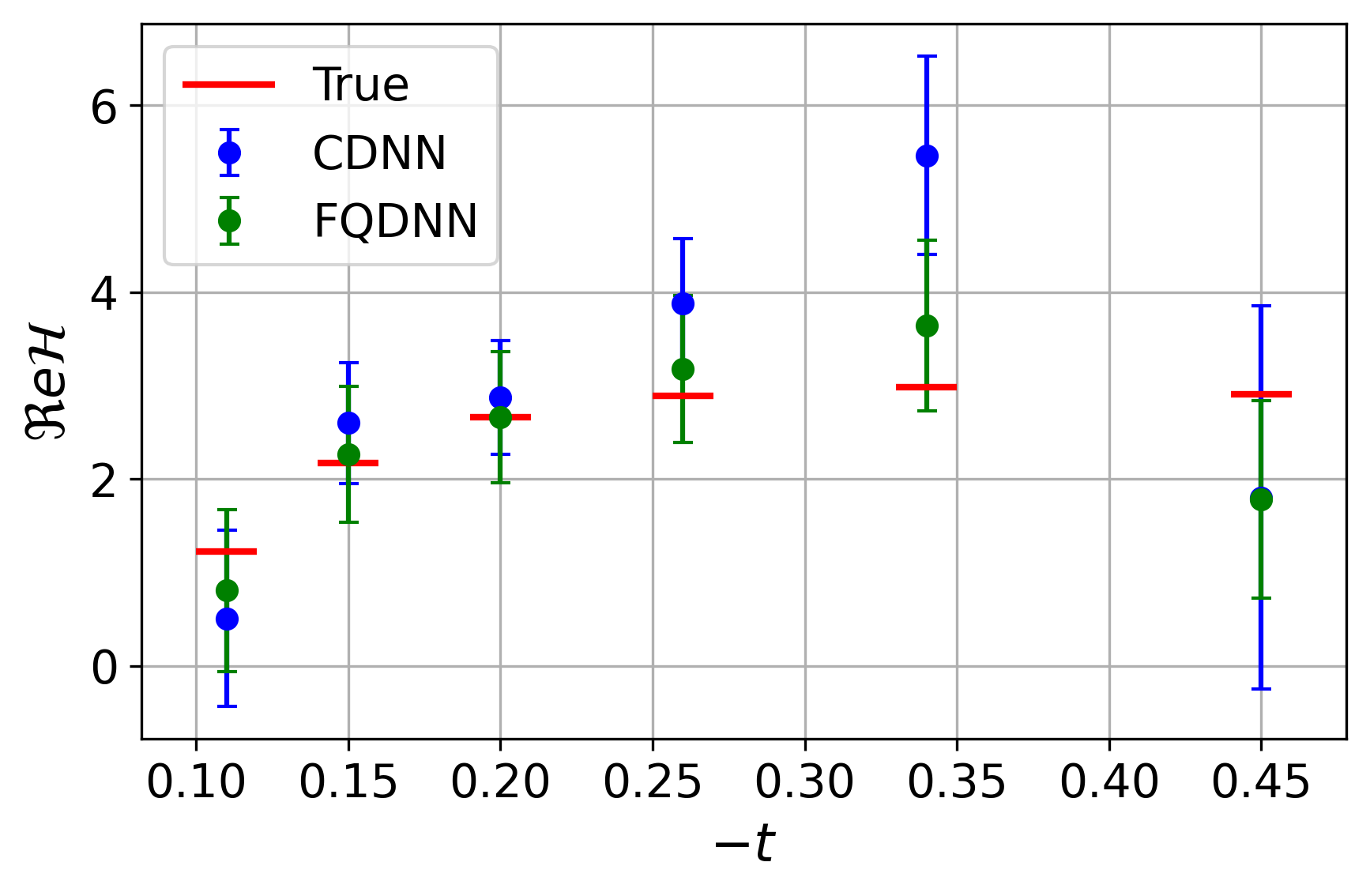}\label{fig:iter-pseudodata-ReH}} \\
    \subfloat[$\Re e\mathcal{E}$ vs $-t$ with $x_B=0.304$]{\includegraphics[width=0.375\textwidth]{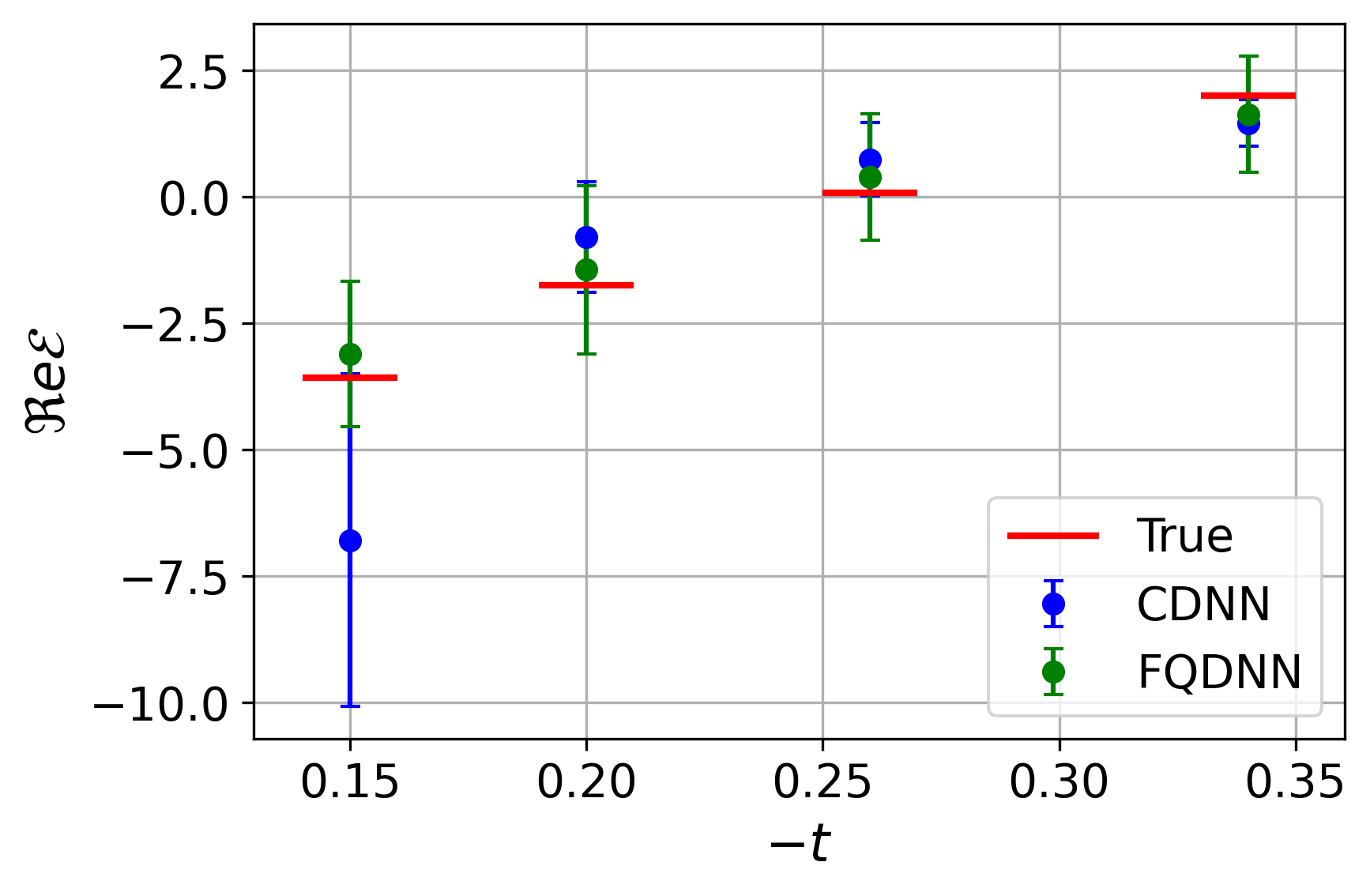}\label{fig:iter-pseudodata-ReE}} \\
    \subfloat[$\Re e\widetilde{\mathcal{H}}$ vs $-t$ with $x_B=0.399$]{\includegraphics[width=0.375\textwidth]{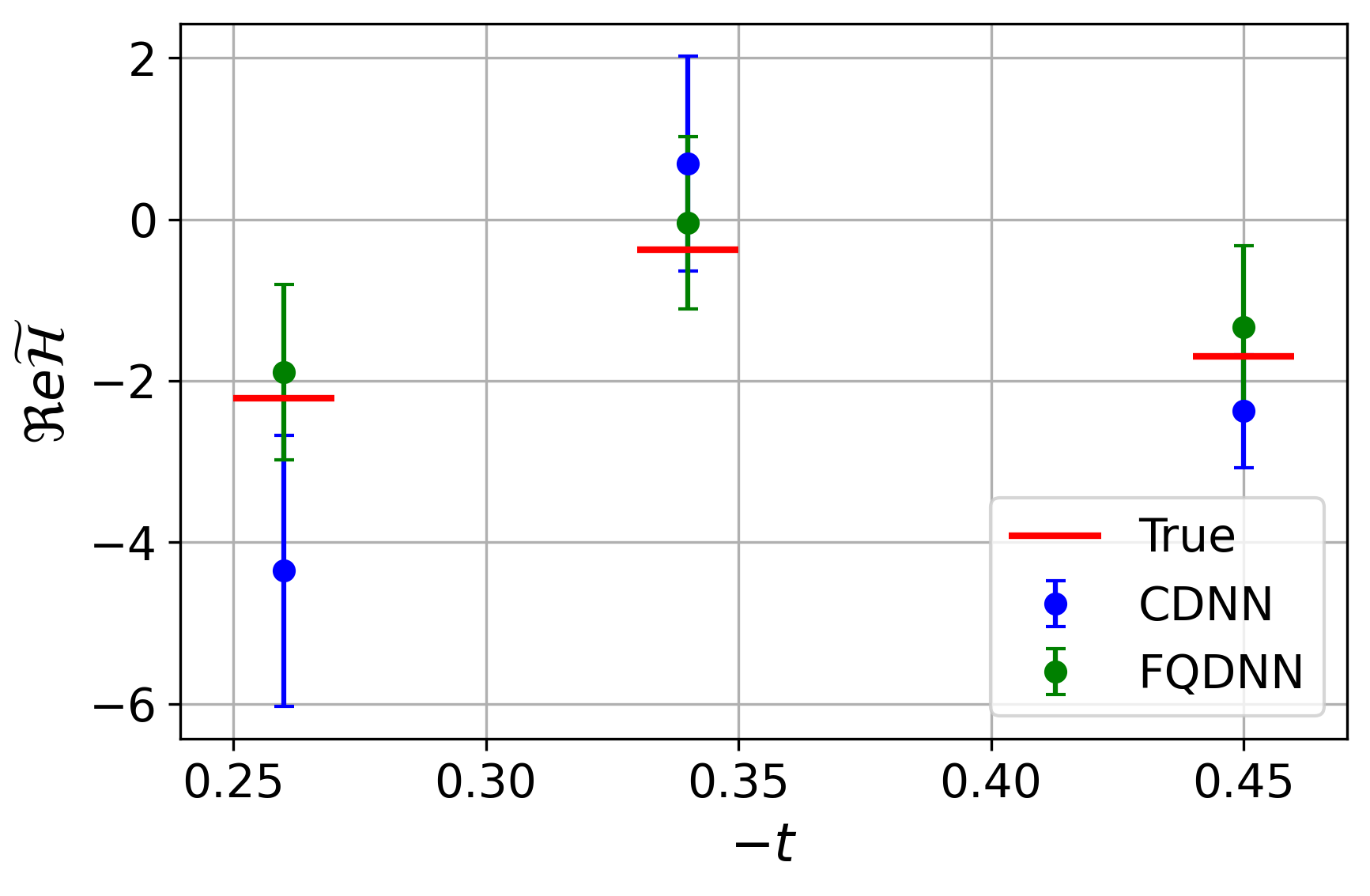}\label{fig:iter-pseudodata-ReHt}} \\
    \subfloat[$DVCS$ vs $-t$ with $x_B=0.335$]{\includegraphics[width=0.375\textwidth]{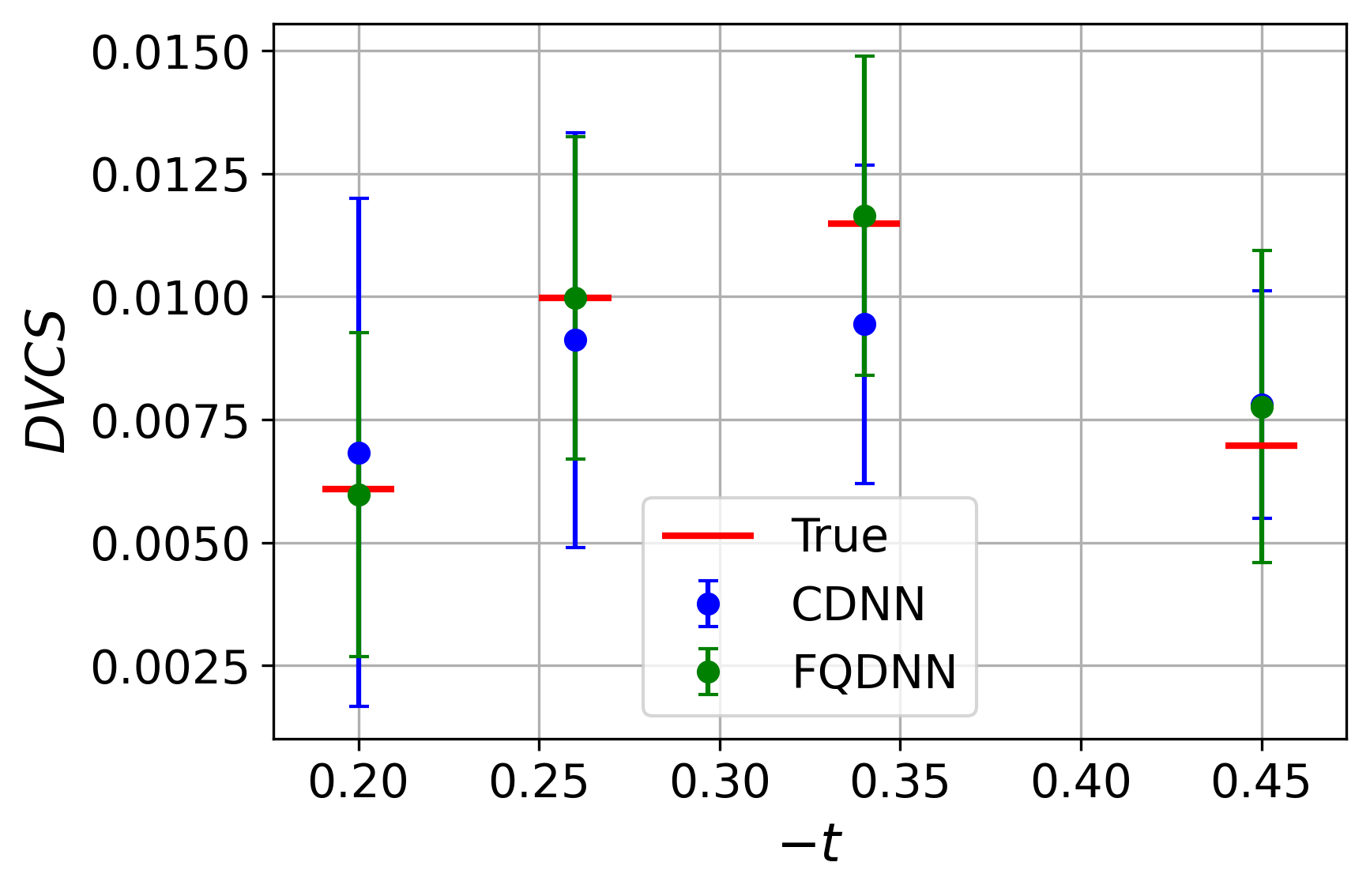}\label{fig:iter-pseudodata-dvcs}} 
    \caption{Example comparisons of CFFs extracted by the tuned CDNN and FQDNN from realistic pseudodata. The CDNN results are shown in blue, the FQDNN results are shown in green, and the true CFFs are shown as red lines. These plots reflect the improved FQDNN performance in terms of accuracy and precision as predicted by the DVCS quantum qualifier.\label{fig:iter-pseudodata-res}}
\end{figure}
\noindent

\begin{figure}[htbp!]
    \centering
    \subfloat[$\Re e\mathcal{H}$ vs. $x_B$ and $-t$]{\includegraphics[width=0.4\textwidth]{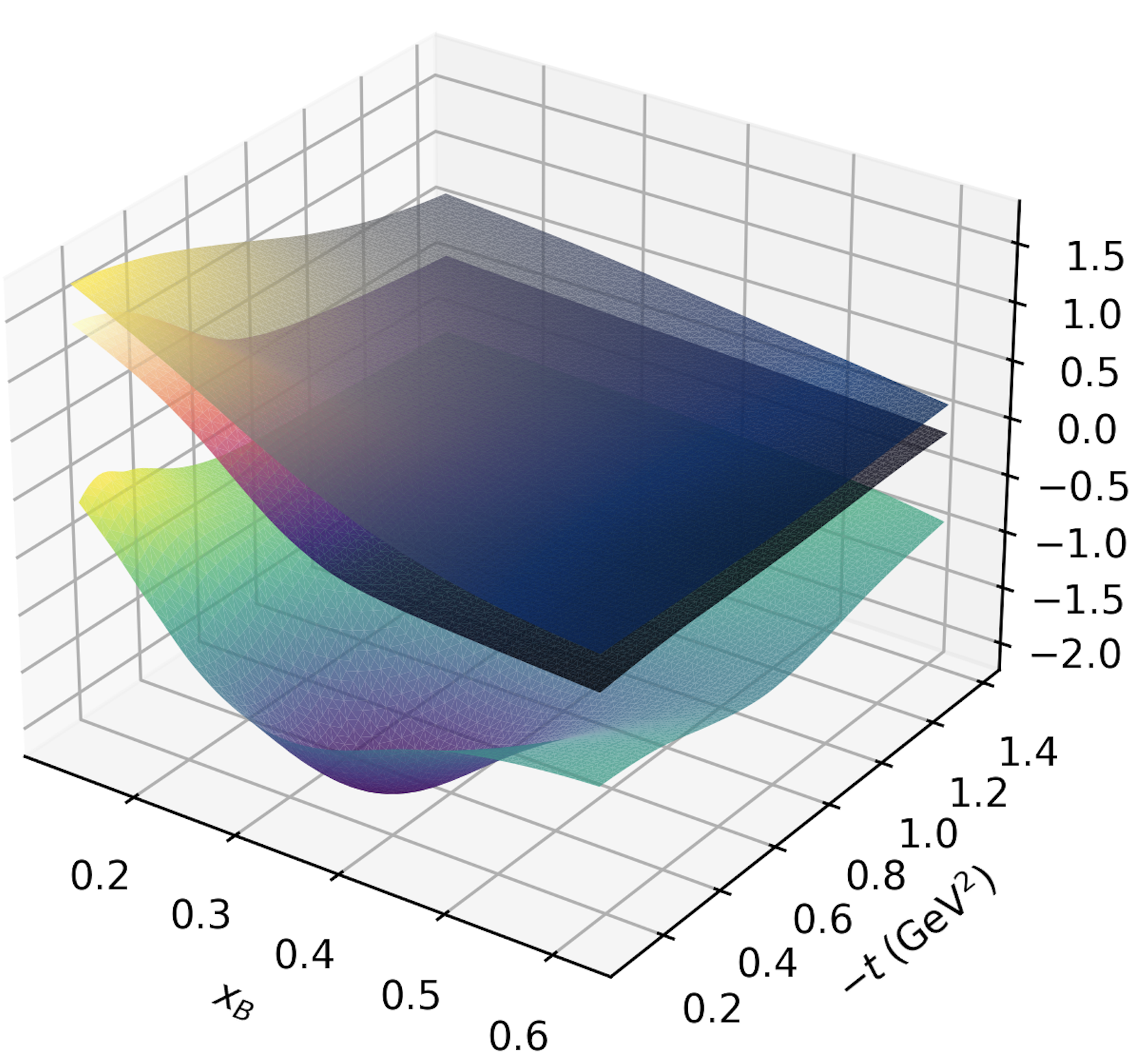}} \\
    \subfloat[$\Re e\mathcal{E}$ vs. $x_B$ and $-t$]{\includegraphics[width=0.4\textwidth]{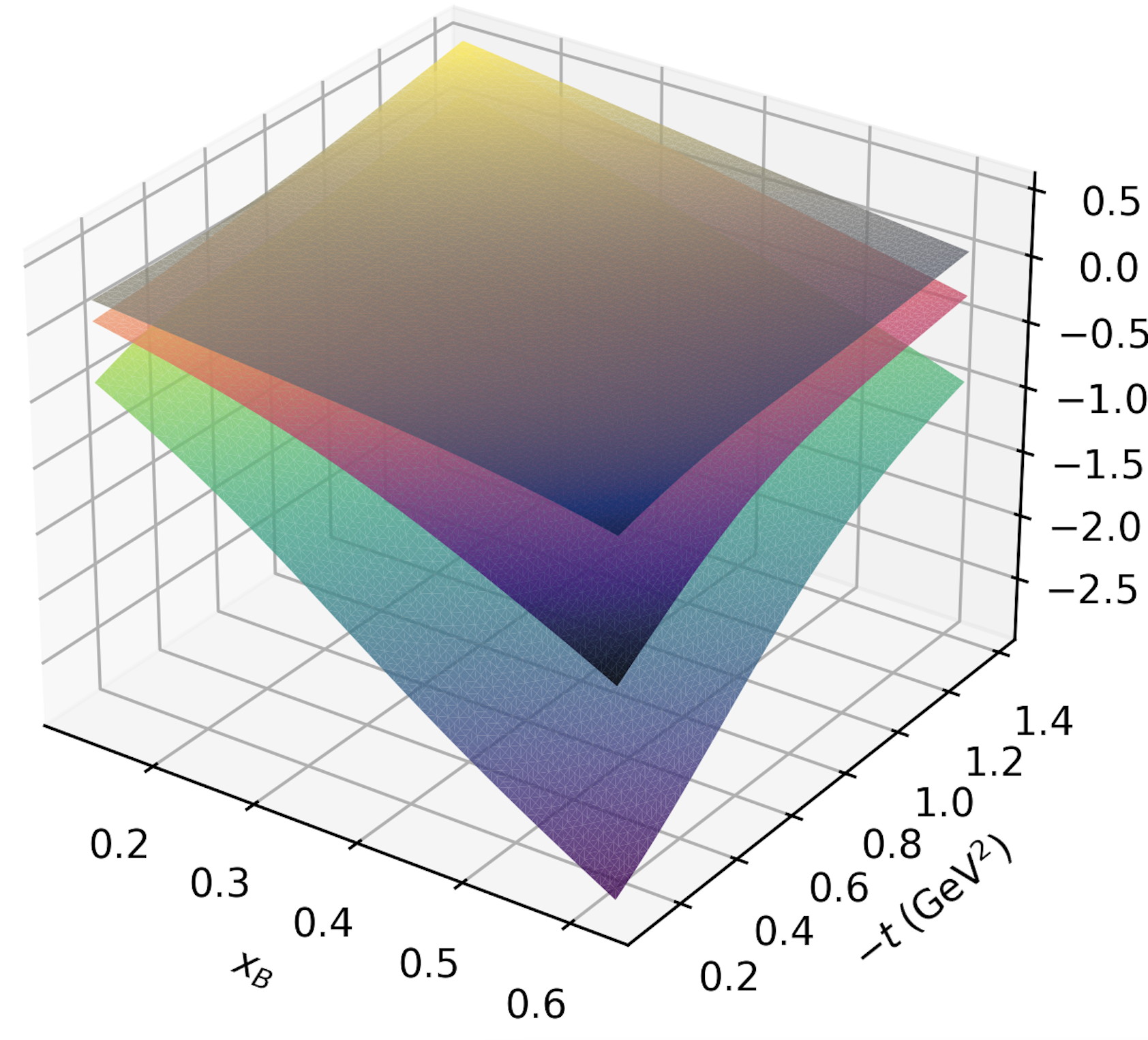}} \\
    \subfloat[$\Re e\widetilde{\mathcal{H}}$ vs. $x_B$ and $-t$]{\includegraphics[width=0.4\textwidth]{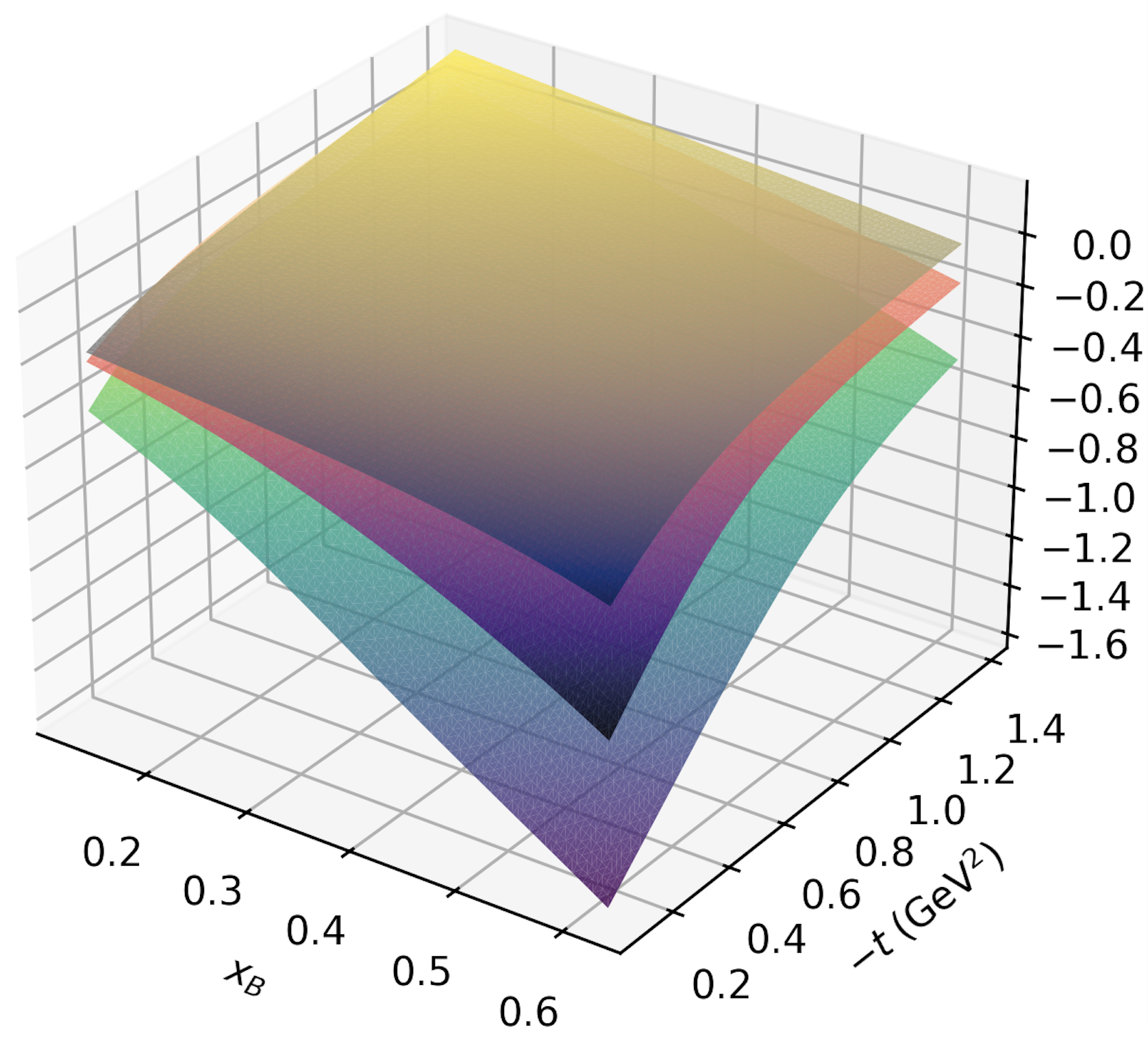}\label{fig:3D-ReHt}} \\
    \caption{Three-dimensional surface plots of (a) $\Re e\mathcal{H}$, (b) $\Re e\mathcal{E}$, (c) $\Re e\widetilde{\mathcal{H}}$ vs. $x_B$ and $-t$, generated showing the mean of the global fit result for three values of $Q^2$ 1.0, 4.0, 8.0 GeV$^2$ with 1.0 GeV$^2$ on the bottom in each case. \label{fig:iter-real-3d}}
\end{figure}

\subsection{Realistic Pseudodata Extraction}
\label{subsec:realistic-pseudo-extract}

We perform both CDNN and FQDNN extractions on the realistic pseudodata sets to optimize the performance of the deep neural networks for application to real experimental measurements. This optimization is achieved by tuning key hyperparameters of each network. For both models, the number of training epochs and the learning rate are varied, while for the FQDNN specifically, additional quantum-specific parameters—such as entanglement strength and input normalization—are systematically scanned. These hyperparameters are adjusted to maximize both the accuracy, defined as the agreement with the true CFFs used to generate the pseudodata, and the precision, defined as the stability of the extracted CFFs across noisy replica datasets.

The quantum qualifier DVCS $\hat{\Xi}_{DVCS}$ [Eq.~\eqref{eq:dvcs-qq}] indicates that the FQDNN should outperform nearly all ($96\%$) of the pseudodata sets. This is confirmed by the realistic pseudodata extraction, which shows overwhelming quantum outperformance. In Fig.~\ref{fig:iter-pseudodata-res}, we plot example comparisons of the extracted CFFs by the tuned CDNN and FQDNN. As predicted by the DVCS quantum qualifier, the FQDNN demonstrates improved accuracy and precision across the entire range of kinematic values covered by the realistic pseudodata sets, effectively capturing the trend in the CFFs.

\begin{figure}[htbp!]
    \centering
    \subfloat[$\Re e\mathcal{H}$ vs $-t$ with $x_B=0.365$]{\includegraphics[width=0.45\textwidth]{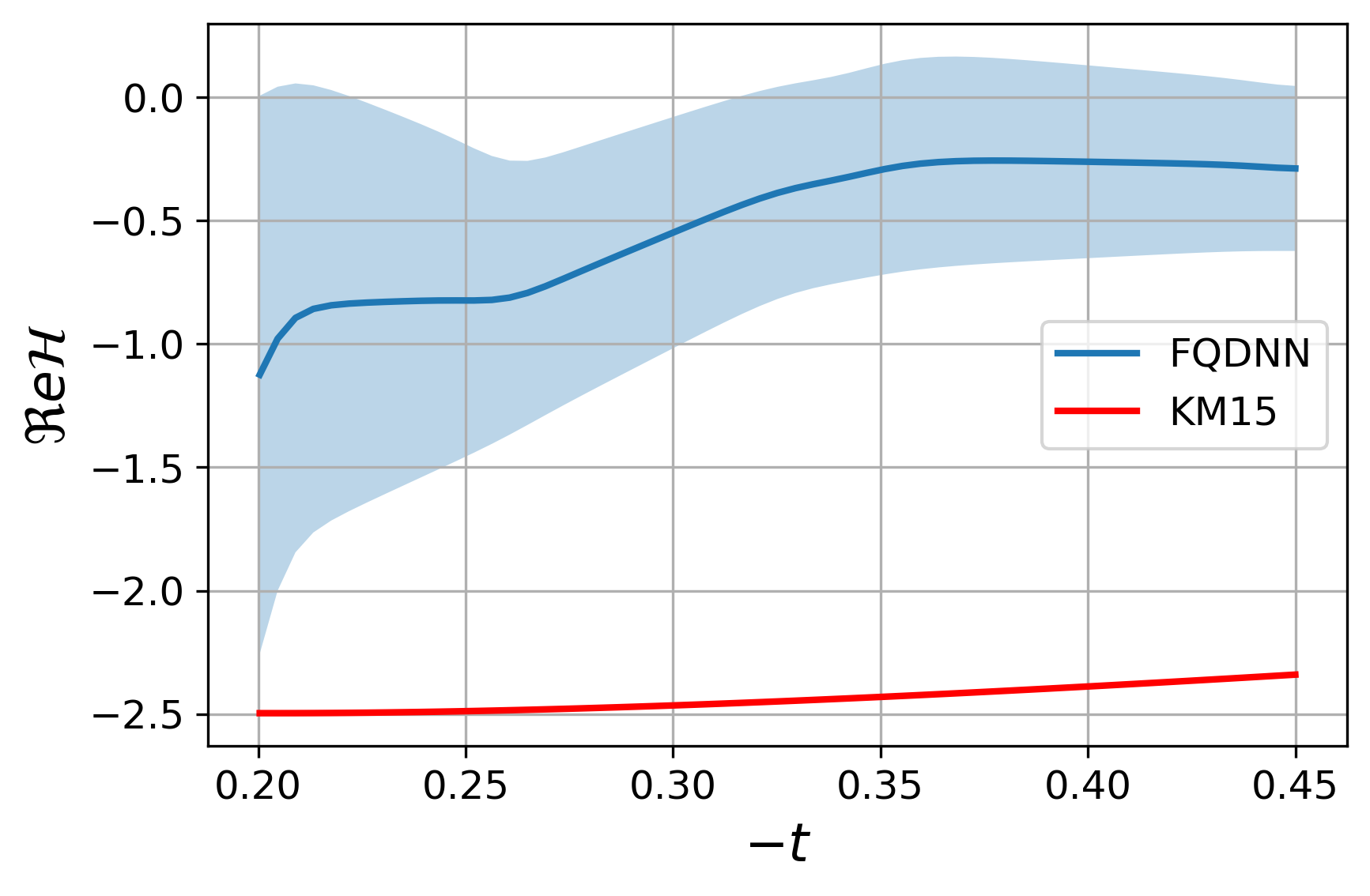}\label{fig:iter-real-ReH}} \\
    \subfloat[$\Re e\mathcal{E}$ vs $-t$ with $x_B=0.275$]{\includegraphics[width=0.45\textwidth]{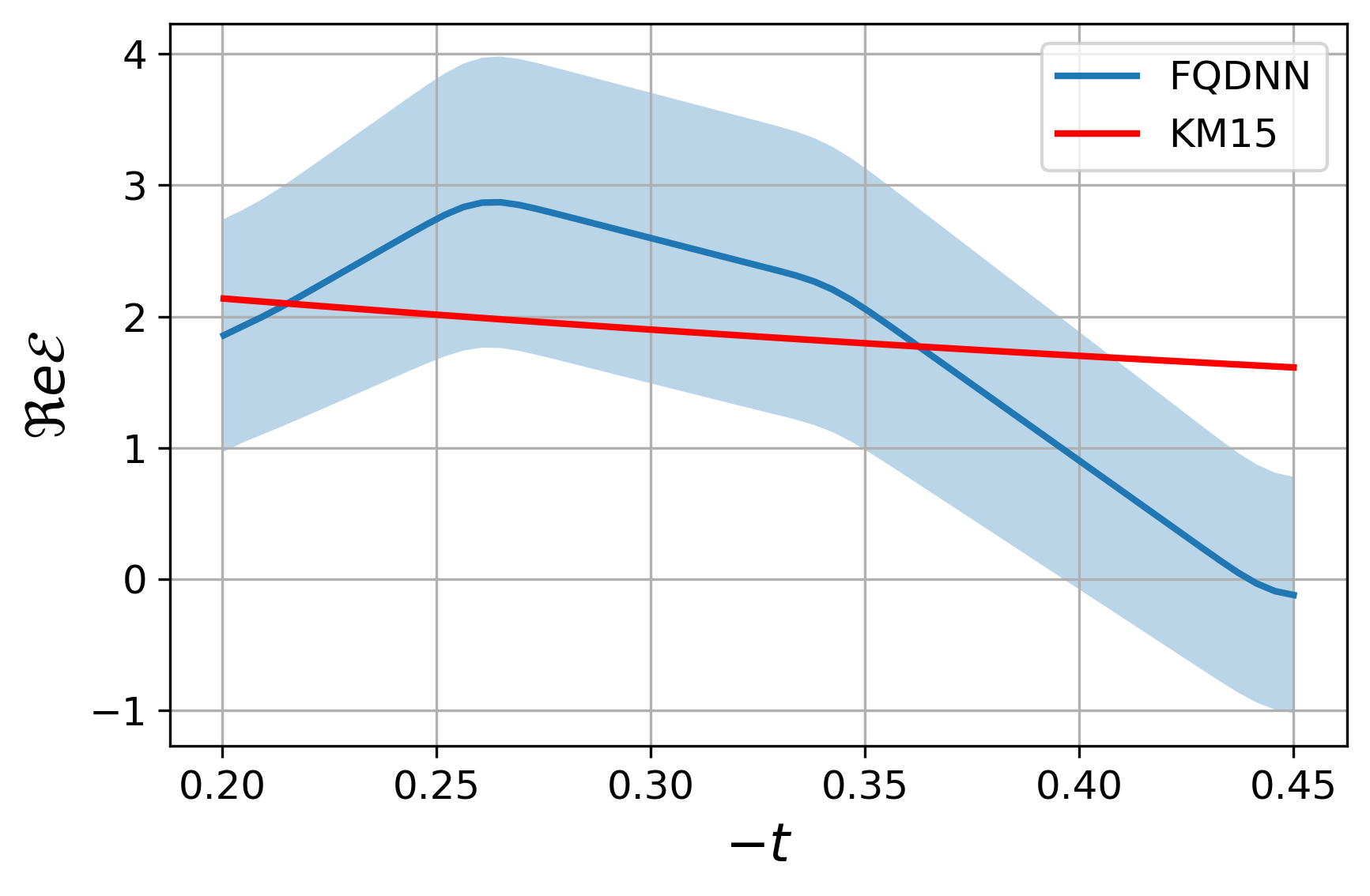}\label{fig:iter-real-ReE}} \\
    \subfloat[$\Re e\widetilde{\mathcal{H}}$ vs $-t$ with $x_B=0.305$]{\includegraphics[width=0.45\textwidth]{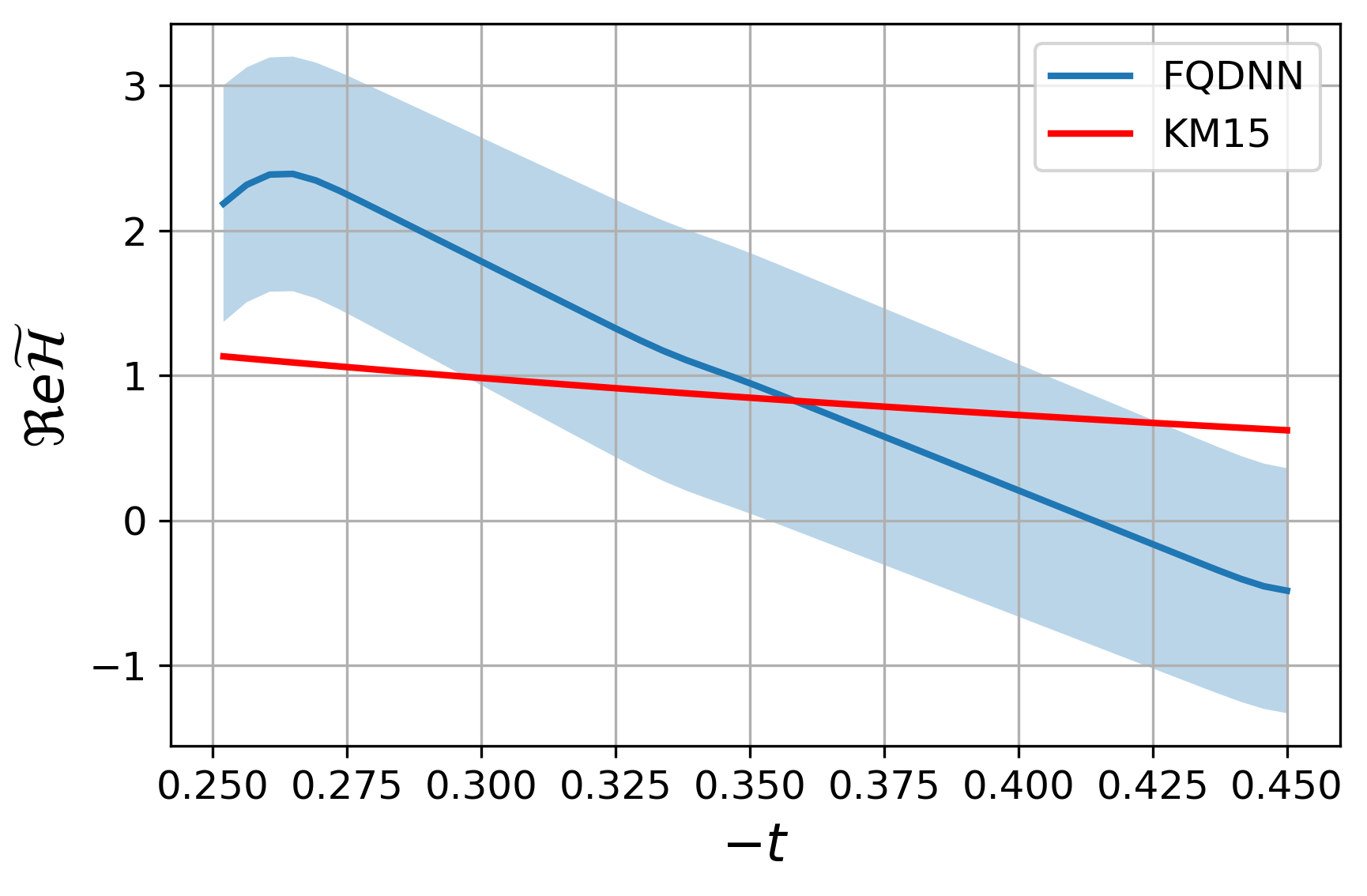}\label{fig:iter-real-ReHt}} \\
    \caption{Example comparisons between the FQDNN-extracted CFFs from the real experimental data and the KM15 prediction. The FQDNN mean is show as the blue line with shaded $1\sigma$ error band, and the KM15 curve is shown in red. }
    \label{fig:iter-real-examples}
\end{figure}

\subsection{Experimental Data Extraction}
\label{subsec:exp-data-extract}
The construction of $\hat{\Xi}_{\text{DVCS}}$ allows us to go back and evaluate the entire experimental dataset in Table \ref{tab:data_summary} and fit each local dataset using either the FQDNN or the standard CDNN for best results.  We find that over 60\% of the data benefits from the FQDNN approach.  Roughly 20\% of the data fits equally well with either algorithm.  The remaining 20\% fits better with the CDNN approach.  All of the optimally fitted CFFs (either with FQDNN or CDNN) are subsequently used in a standard DNN-based global fit to obtain a continuous CFF model throughout the measured phase space.

The global fit follows exactly the same procedure as in our previous $\chi$DNN analysis~\cite{Diaz:2025}. For $\Re e\mathcal{H}$ (and analogously for $\Re e\mathcal{E}$ and $\Re e\widetilde{\mathcal{H}}$), the DNN takes input variables (kinematics plus skewness-dependent quantities) and employs a deep feed-forward architecture with 8 hidden layers containing 36 neurons per layer. A mixture of ReLU, Leaky ReLU, tanh, ReLU6, and tanhshrink activations is used, combined with batch normalization and dropout regularization (rates 0.1--0.5) to ensure stability and prevent over-fitting.

Experimental and algorithmic uncertainties are propagated simultaneously by training 1000 independent DNNs on bootstrap replicas of the data set. At any kinematic point $(x_B,t,Q^2)$, the central CFF value is taken as the ensemble mean $\overline{\mathcal{F}}_\text{DNN}$, and the total uncertainty is given by the standard deviation of the replica predictions,
\begin{equation}
\sigma(x_B,t,Q^2) =
\sqrt{\frac{1}{N-1}\sum_{j=1}^{N}
\left(\mathcal{F}^j_{\text{DNN}} - \overline{\mathcal{F}}_{\text{DNN}}\right)^2},
\end{equation}
with $N=1000$. The resulting continuous CFF surfaces are shown as functions of $x_B$ and $-t$ in Fig.~\ref{fig:iter-real-3d}.

For consistency checks, we compare our resulting global fit to the KM15 model \cite{Kumericki2015}, a modern global fit for DVCS that parametrizes the dominant CFFs using a flexible GPD ansatz. It builds on earlier KM10b \cite{Kumericki2010} and KMM12b \cite{Kumericki2012} fits and incorporates updated DVCS cross section and asymmetry data, including the earlier, but incomplete data of same measurements from Hall A E07-007 and E00-110  \cite{Camacho2006} and CLAS Hall B e1-DVCS1 \cite{Girod2008,Gavalian2009} as in Table \ref{tab:data_summary}. The model includes both $H$ and $E$ GPDs, with a simplified functional form that allows for control over $t$-dependence and skewness effects. It assumes factorized $t$-dependence and parametrizes the GPDs using Regge-inspired profiles and a spectral representation consistent with polynomiality constraints. The KM15 fit reveals that the skewness effect in GPD $H$ is small, and the $t$-dependence of $E$ is flatter than that of $H$, aligning with expectations from dispersion relations. The model is constrained by experimental observables through Fourier harmonics of the cross section, with sensitivity enhanced by weighting techniques.  Fig.~\ref{fig:iter-real-examples} shows that the our global replica mean as the blue line is consistent within $1\sigma$ error bands with the KM15 model (red line) for $\Re e\mathcal{E}$, and $\Re e\widetilde{\mathcal{H}}$, but deviates with a clear increasing trend in $-t$ for $\Re e\mathcal{H}$.

Differences between nominally twist-2 cross-section representations can arise from kinematical power corrections of order $\mathcal{O}(M^2/Q^2)$ or $\mathcal{O}(|t|/Q^2)$, from convention-dependent treatments of subleading terms, and from truncation of twist-3 contributions. Because our forward model is restricted to twist-2 accuracy, any residual higher-twist effects in the data are absorbed into the fitted parameters.
To assess the impact of such effects, we repeated the extraction both with and without kinematic restrictions on $Q^2$ and $|t|/Q^2$. The resulting changes in the extracted CFFs were on average below $3\%$, consistent with replica-level statistical fluctuations. We therefore retain the kinematic restrictions in the final analysis to suppress power-suppressed contributions.

\subsection{Comparison to Previous Extractions}
\begin{figure}[htbp!]
\centering
\subfloat[Global DNN predictions for $\Re e\mathcal{H}$ versus $\xi$ at fixed $Q^2 = 2~\mathrm{GeV}^2$ and $t = -0.3~\mathrm{GeV}^2$ in comparison with results from Mou19 \cite{Moutarde:ANN2019} and $\chi$DNN \cite{Diaz:2025} along with KM15 \cite{Kumericki2015}.]{\includegraphics[width=0.45\textwidth]{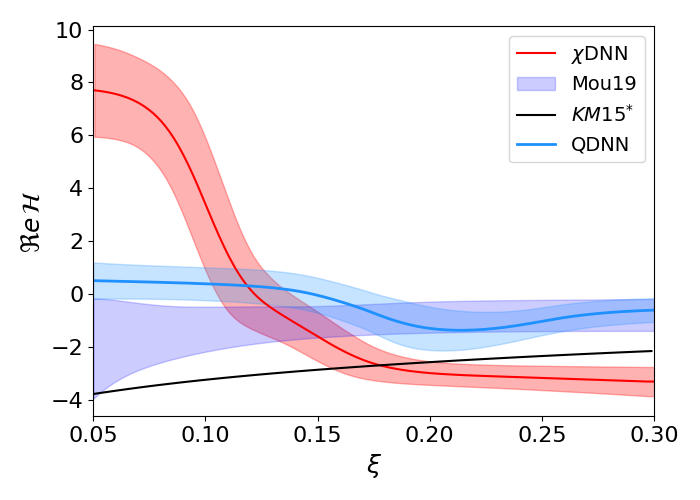}\label{fig:comparison1}} \\
\subfloat[Global DNN predictions for $\Re e\mathcal{E}$ versus $\xi$ at fixed $Q^2 = 4~\mathrm{GeV}^2$ and $t = -0.2~\mathrm{GeV}^2$ in comparison with results from NN20 and NNDR20 \cite{kumericki:2020} and $\chi$DNN \cite{Diaz:2025} along with KM15 \cite{Kumericki2015}.]{\includegraphics[width=0.45\textwidth]{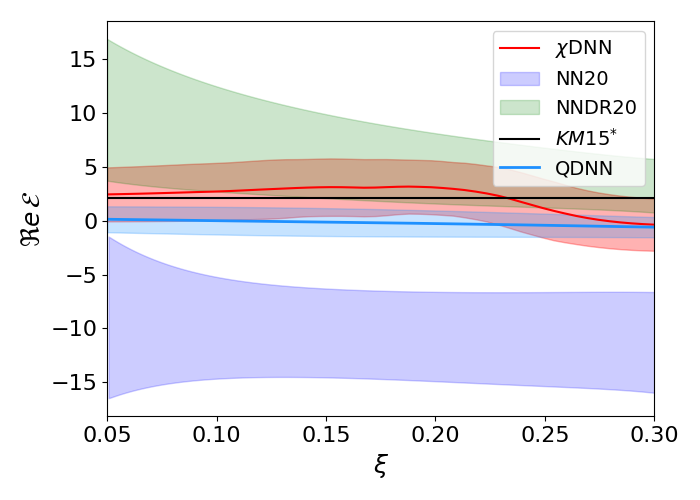}\label{fig:comparison2}} \\
\subfloat[Global DNN predictions for $\Re e\widetilde{\mathcal{H}}$ versus $\xi$ at fixed $Q^2 = 2~\mathrm{GeV}^2$ and $t = -0.3~\mathrm{GeV}^2$ in comparison with results from \cite{Moutarde:ANN2019} and $\chi$DNN \cite{Diaz:2025} along with KM15 \cite{Kumericki2015}.]{\includegraphics[width=0.45\textwidth]{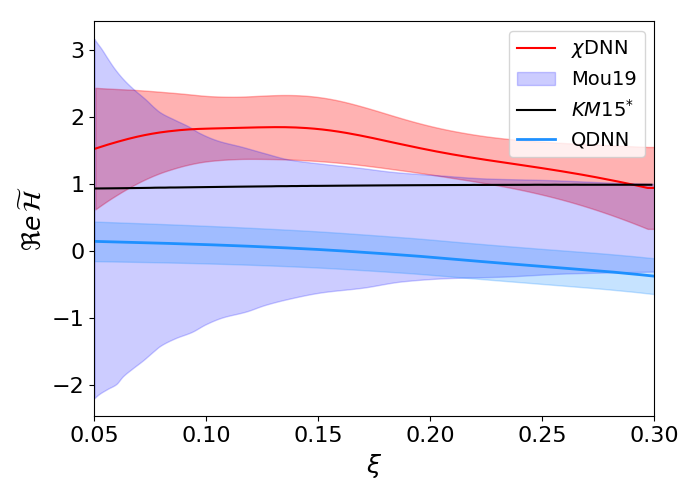}\label{fig:comparison3}} 
\caption{Comparison of our global DNN predictions with previous extractions from experimental data.}
\label{fig:comparison-fit}
\end{figure}
Previous global analyses based on classical deep neural networks established that such approaches can provide an effective, model-independent means of extracting CFFs from DVCS observables. Most of those modern studies (besides $\chi$DNN \cite{Diaz:2025}) combined unpolarized cross sections with beam-spin asymmetries to constrain some parts of both real and imaginary components of the CFFs across a broad kinematic range. While such global DNN fits represented a major step beyond traditional parameterizations, their performance remained closely tied to the number and observables used to constrain the fit, and they often required auxiliary physical constraints or dispersion-relation regularization to ensure convergence.  The $\chi$DNN \cite{Diaz:2025} and our results here fit use only the unpolarized cross section, but also constrain the parameterization to only the four parameters $\Re e\,\mathcal{H}$, $\Re e\,\mathcal{E}$, and $\Re e\,\widetilde{\mathcal{H}}$, and the DVCS term, as described in Section \ref{sec:TheoryFramework}.

Unique to the present analysis, the QDNN approach adopts a fundamentally different training hierarchy resulting in generally narrow models (higher precision) overall as seen in the comparison plots in Fig.~\ref{fig:comparison-fit}. The QDNN encodes the response of each kinematic bin in a quantum-inspired latent space that preserves amplitude-level correlations. These locally constrained CFFs are then supplied to a conventional global DNN to reconstruct the full functional dependence. In the case of $\chi$DNN \cite{Diaz:2025}, the same exact data was used to achieve the local fits as was used for the present work. In addition, nearly identical architectures were used in the global phase of the fits.  In this regard, the comparison of these two models in particular constitutes a true like-for-like benchmark: the distinction lies entirely in how the local information is obtained.

The improvement introduced by the QDNN manifests itself not merely as a narrower uncertainty band but as a qualitatively different behavior of the uncertainty itself. Regions of limited experimental coverage show expected widening, yet elsewhere the uncertainties shrink markedly—reflecting the ability of QDNN to exploit latent correlations to stabilize the learning of 
$\Re e\,\mathcal{H}$, $\Re e\,\mathcal{E}$, and $\Re e\,\widetilde{\mathcal{H}}$. This intrinsic stabilization emerges without imposing external physical constraints, indicating that the quantum-encoded latent space acts as a \emph{self-regularizing prior}.

From a methodological standpoint, the QDNN therefore extends the concept of a physics-informed network beyond classical architectures: it learns the complex-valued structure of the DVCS amplitude directly, rather than inferring it indirectly from nonlinear regressions on the observables. The result is a more stable convergence behavior, improved generalization to unseen kinematics, and a reduction in model bias that persists even when trained on the least-constraining dataset.

Overall, the QDNN approach demonstrates that quantum-differentiable representations can reproduce—and in several cases exceed—the precision of prior DNN global fits that relied on a richer set of observables. This establishes the QDNN as a robust and scalable framework for future global extractions, capable of incorporating additional polarization observables without sacrificing interpretability or stability.

\section{\label{sec:con}Conclusion}

Within a common BKM10 twist-2 forward model, we have presented the first extraction
of Compton form factors (CFFs) from experimental DVCS cross sections using a
quantum deep neural network (QDNN), together with a controlled, like-for-like
benchmark against a classical deep neural network (CDNN) for this application.
Across extensive closure tests on kinematically realistic pseudodata and
validation on experimental subsets, optimized QDNNs achieve tighter CFF
uncertainties and, in many regimes, comparable or improved accuracy relative to
CDNNs. The gains are most pronounced in sparse or high-uncertainty kinematic
regions and for more complex (more nonlinear) $\phi$-dependence of the measured
cross section.

Methodologically, we identified and mitigated optimization instabilities that are
specific to quantum-circuit models. In particular, tunable entanglement range,
layer-wise depth growth, and depth-scaled initialization stabilize training and
reduce algorithmic variability, bringing algorithmic errors to a level comparable
to classical baselines while producing visibly tighter uncertainty bands in the
reconstructed cross sections.

A practical outcome of this study is the DVCS quantum qualifier
$\hat{\Xi}_{\mathrm{DVCS}}$: a simple, data-driven selection rule built from
measurable features of a dataset (the average scaled experimental error and a
nonlinearity metric) that predicts \emph{a priori} when a QDNN is expected to
outperform a CDNN. In our tests, the qualifier successfully guided per-bin model
selection and enabled an optimized strategy in which the best-performing local
extraction (QDNN or CDNN) is retained for each kinematic bin.

Applying this qualifier-guided approach to the full set of experimental data yields a set of
high-quality local CFF extractions that we then use to train a global DNN
parameterization over the measured phase space. The resulting trends are
consistent with established global analyses for $\Re e\,\mathcal{E}$ and
$\Re e\,\widetilde{\mathcal{H}}$ within uncertainties, while $\Re e\,\mathcal{H}$
exhibits a systematic difference in its $t$-dependence. Overall, the
combined local and global strategy delivers consistently reduced uncertainties across
the kinematic range compared to representative previous extractions.

Looking ahead, the framework is naturally extensible to full eight-CFF and
polarization-dependent extractions, and to increased theory fidelity through QCD
evolution and next-to-leading-order effects. Although the present results are
obtained with simulator-based quantum circuits, the architectures are
hardware-compatible, providing a direct path to native execution as quantum
processors mature. The framework and software used in this analysis are
available at the UVA Spin Physics GitHub~\cite{uva2025qdnn}.

\section*{Acknowledgments}

The Deep Neural Network models used in this work were trained using the University of Virginia’s high-performance computing cluster, Rivanna.
The authors acknowledge Research Computing at the University of Virginia for providing computational resources and technical support that have contributed to the results reported in this publication. URL: \url{https://rc.virginia.edu}.
This work was supported by the DOE contract DE-FG02-96ER40950.


%

\end{document}